\documentclass{article} % For LaTeX2e
\usepackage{iclr2027_conference,times}

\usepackage{amsmath,amsfonts,bm}

\def\eqref#1{equation~\ref{#1}}
\def\1{\bm{1}}

\DeclareMathAlphabet{\mathsfit}{\encodingdefault}{\sfdefault}{m}{sl}
\SetMathAlphabet{\mathsfit}{bold}{\encodingdefault}{\sfdefault}{bx}{n}

\usepackage[most]{tcolorbox,adjustbox}

\newtcolorbox{promptbox}{
      enhanced,
      breakable,
      colback=black!4,
      colframe=black!18,
      boxrule=0.4pt,
      arc=2pt,
      left=7pt,
      right=7pt,
      top=6pt,
      bottom=6pt,
      before skip=5pt,
      after skip=8pt,
      fontupper=\small\ttfamily,
}

\usepackage{hyperref}
\usepackage{url,booktabs,amssymb,wrapfig}
\usepackage{graphicx}
\usepackage{caption}
\usepackage{subcaption}
\usepackage{algorithm,algorithmic,multirow}
\usepackage{autobreak}
\allowdisplaybreaks

\usepackage{pifont}
\newcommand{\cmark}{\ding{51}}
\newcommand{\xmark}{\ding{55}}

\title{VisAudit: Evaluating  Multimodal Agents for Visual Diagnosis and Repair}

\iclrfinalcopy
\author{
Shicheng Liu \And
Adam Kahirov \And
Qi Zhang \And
Zhimin Hu \AND
Song Wang \And
Junhong Lin \And
Julian Shun \And
Yada Zhu
}

\begin{document}

\maketitle

\begin{abstract}
Multimodal agents are increasingly used for data visualization tasks but remain limited in autonomous review. Unlike humans, they may fail to recognize when a visualization is incorrect, determine what to change, repair it without disrupting correct content, and verify whether the intervention succeeded. Existing benchmarks largely evaluate predefined individual capabilities such as chart generation, instruction-guided editing, or defect detection, and therefore do not capture this gap in autonomous review. We introduce VisAudit, a benchmark for evaluating visualization diagnosis, repair, and verification. Given a rendered chart and configurable auxiliary evidence, including its source data table, intended text summary, and visualization code, an agent iteratively diagnoses potential defects, modifies and executes visualization code, inspects execution and visual feedback, and determines when no further intervention is needed. VisAudit defines three tracks spanning diagnosed repair, autonomous repair, and open-world verification, and contains 1,900 flawed instances across 21 chart types and 10 flaw categories, together with 300 initially correct charts. We construct the benchmark through controlled perturbations of validated source visualizations, with systematic verification and human-aligned quality control to ensure that injected defects are well-defined and recoverable from the available evidence. Experiments with leading multimodal models reveal a substantial gap from reliable autonomous review: the strongest evaluated model fully recovers only $47.4\%$ of flawed charts in the autonomous-repair setting. By exposing failures across diagnosis, intervention, preservation, and verification, VisAudit characterizes the capabilities needed to close the gap between current agents and human-like visualization review and provides actionable insights for designing more reliable autonomous multimodal systems.
\end{abstract}

% \zhu{I think the flow of the abstact can be improved. For example, "we construct the benchmark..." should be moved before "Experiments with..." You describe the benchmark content and construction, then experiment results, then finally the value of the benchmark and results.}

%Data visualizations are widely used to communicate and reason about data, but construction errors can distort the information they convey. Existing benchmarks primarily evaluate chart generation, instruction-guided editing, or defect detection, leaving open whether multimodal agents can autonomously review visualizations. We introduce VisAudit, a benchmark for autonomous visualization diagnosis, repair, and verification. Given a chart and configurable auxiliary evidence, an agent iteratively diagnoses potential defects, repairs and executes visualization code, inspects the resulting feedback, and decides when no further intervention is needed. VisAudit contains 1,900 flawed instances across 21 chart types and 10 flaw categories, together with 300 correct charts, and defines three tracks spanning diagnosed repair, autonomous repair, and open-world verification. Experiments with leading multimodal models reveal substantial limitations in autonomous review, with the strongest evaluated model fully recovering only $39.65\%$ of flawed charts in the autonomous-repair setting. By exposing where agents fail across diagnosis, intervention, and verification, VisAudit provides a testbed for advancing the next generation of autonomous multimodal agents.

\section{Introduction}

Multimodal agents are increasingly capable of generating, editing, and reasoning about data visualizations \citep{yang2025chartmimic,zhao2025chartedit,tang2026charts,rahman2026viseditbench}. However, autonomous visualization review requires more than executing a predefined task. When humans review a visualization, they can determine whether anything is wrong, identify the underlying problem, decide whether and how to intervene, and verify whether the result is correct. An autonomous agent should similarly be able to discover unspecified defects, determine appropriate interventions, preserve content that is already correct, and recognize when no further action is necessary. However, autonomous recognition and correction of failures remains challenging for current tool-using and multimodal agents \citep{liu2026llms,dixit2026evaluating}. This raises a fundamental question: \emph{Can an agent, like a human reviewer, autonomously determine when intervention is needed, decide what to do, and verify that its intervention succeeded?}

%This workflow requires capabilities beyond instruction following. When humans review an artifact, they can inspect the available evidence, identify potential problems, decide whether intervention is warranted, perform a correction, and examine the result to determine whether the problem has been resolved. An autonomous agent must similarly discover problems without being told what to change, determine whether and how to intervene, preserve correct content, and verify the consequences of its actions. Yet these capabilities remain challenging for current agents: recent studies of tool-using and multimodal agents report substantial gaps between task execution and autonomous recognition and correction of failures \citep{liu2026llms,dixit2026evaluating}. Autonomous visualization review therefore provides a concrete setting for a broader question in agentic AI: \emph{Can an agent, like a human reviewer, determine when intervention is needed, decide what to do, and verify that its intervention succeeded?}

Existing visualization benchmarks do not capture this form of autonomy. Recent benchmarks evaluate chart generation and reconstruction \citep{yang2025chartmimic,tang2026charts}, instruction-guided editing \citep{zhao2025chartedit,kapadnis2026charteditbench,zhang2026realchart2code,rahman2026viseditbench}, or misleading-visualization detection \citep{chen2025unmasking,tonglet2026chart}. Despite their different formulations, these benchmarks generally define the task to be performed in advance: the agent is asked to generate a chart, execute a specified modification, or assess a particular property of the visualization. In realistic review, however, the task itself may not be predetermined. Given a visualization, an agent must autonomously determine whether any problem exists, what intervention, if any, is warranted, and whether its actions have produced a satisfactory result. This shifts the requirement from completing a predefined visualization task to autonomously managing the full review process, including the possibility that no action should be taken.

\begin{figure}[t]
    \centering
    \includegraphics[width=\linewidth]{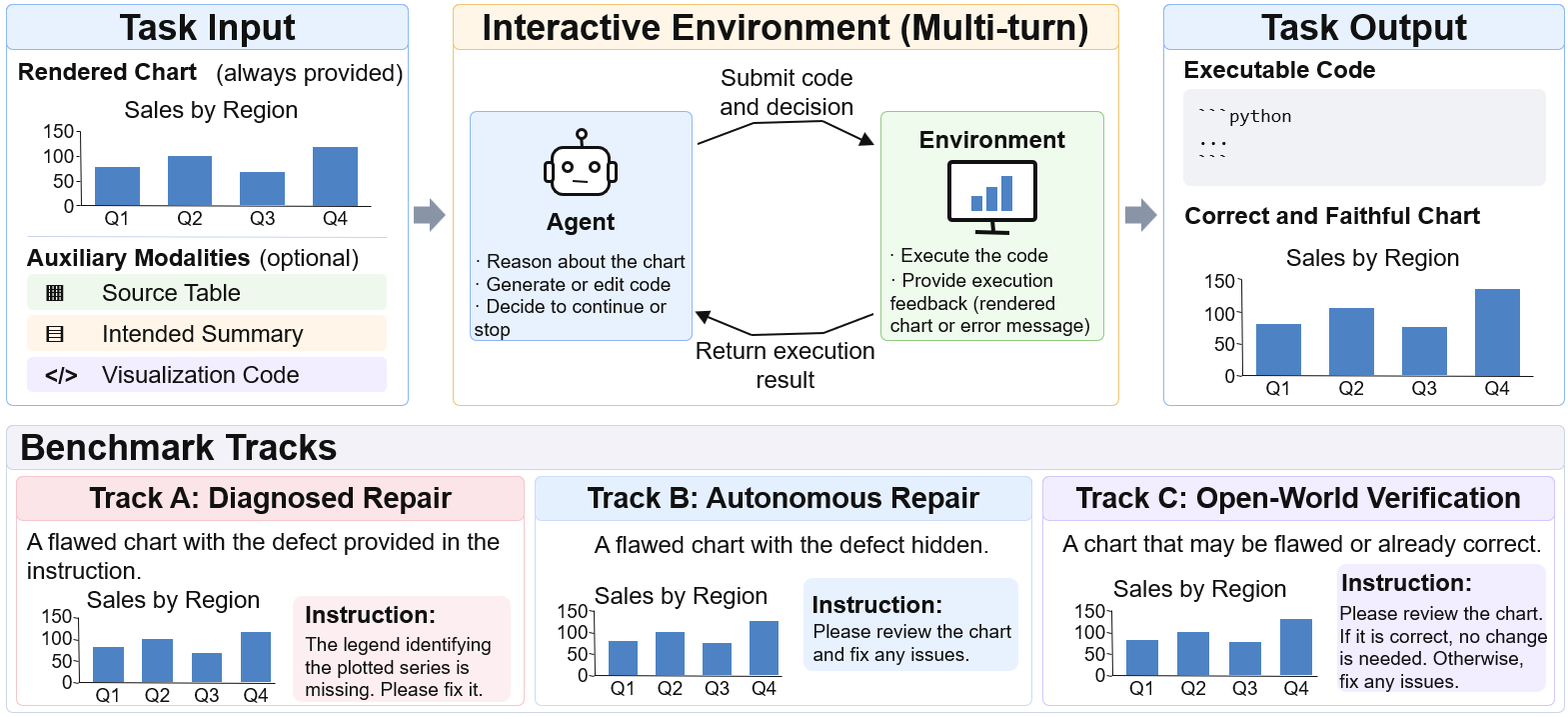}
    \caption{Overview of VisAudit. Given a chart and optional auxiliary modalities, an agent iteratively edits, executes, and verifies visualization code in a multi-turn environment. We evaluate three tracks: (A) diagnosed repair, (B) autonomous repair, and (C) open-world verification.}
    \label{fig:overview}
\end{figure}

% We evaluate this agentic workflow capability through three complementary tracks with
% increasing levels of autonomy. \emph{Track A: Diagnosed Repair} provides the agent with a textual description of the known defect and evaluates its ability to perform the corresponding repair. \emph{Track B: Autonomous Repair}, our primary setting, hides the defect and requires the agent to discover and repair it from the available multimodal evidence. Comparing Tracks A and B separates the difficulty of repair from the additional challenge of autonomous diagnosis. Finally, \emph{Track C: Open-World Verification} contains both flawed and already-correct visualizations. The agent must first determine whether any intervention is necessary: flawed charts require autonomous diagnosis and repair, whereas correct charts should be recognized and left unchanged. Together, the three tracks progress from repair, to diagnosis and repair, to open-world verification, diagnosis, and repair.\zhu{at the end of this section, should we level up the impact of this work? e.g. discover vision language model weakness, provide data for fine tuning, fill in the gap of enhancing agents, ... }

To bridge this gap, we introduce \textbf{VisAudit}, a benchmark for evaluating autonomous visualization diagnosis, repair, and verification. As illustrated in Figure~\ref{fig:overview}, each instance provides an initial chart together with configurable auxiliary modalities, including the source table, intended summary, and visualization code. Given this multimodal evidence, an agent determines whether and how the visualization should be modified and produces executable code that renders a correct and faithful chart. VisAudit places this task in an interactive environment, allowing the agent to propose and execute repairs, inspect the resulting visualization and execution feedback, and revise its solution when necessary. Crucially, the agent decides when to continue or stop, enabling it to test its diagnosis, observe the consequences of its interventions, and verify whether the resulting visualization is satisfactory. %\zhu{in reality, what if there is no summary? would it be helpful to give an example for verification, such as while repairing one chart error, the agent can induce another error.}

We evaluate this workflow through three tracks with increasing levels of autonomy. \emph{Track A: Diagnosed Repair} provides the defect and evaluates the corresponding repair; \emph{Track B: Autonomous Repair} hides the defect, requiring the agent to diagnose and repair it from multimodal evidence; and \emph{Track C: Open-World Verification} further introduces already-correct visualizations, requiring the agent to determine whether intervention is necessary. Together, these tracks isolate increasingly demanding capabilities from repair to autonomous diagnosis and open-world verification. By revealing where current agents succeed and fail across the diagnosis-action-verification loop, VisAudit provides actionable insights into the capabilities needed to develop more reliable and autonomous multimodal agents.

Our main contributions are as follows:

First, we develop a controlled and reusable visual perturbation framework that transforms correct source visualizations into flawed instances, yielding 1,900 flawed instances across 21 chart types and 10 flaw categories, together with 300 initially correct charts for open-world evaluation. 

Second, we introduce three complementary evaluation tracks and a multidimensional evaluation framework that measures executability, recovery, preservation, and correct-chart recognition, capturing both repair effectiveness and the ability to avoid unnecessary interventions.

Third, we extensively evaluate leading multimodal models and conduct modality and interaction ablations, revealing substantial limitations in autonomous visualization review: agents struggle to diagnose unspecified defects, execute some repairs even when the defect is given, completely resolve multiple simultaneous flaws, and avoid unnecessary modifications to already-correct charts.

\section{Related Work}
\begin{table}[H]
\caption{Comparison with representative visualization generation, editing, refinement, and misleading-visualization benchmarks. \textbf{Conserve} denotes explicitly preserving already-correct contents during repair; \textbf{Verify} denotes using the consequence of the model's own repair (e.g., execution and re-rendering) to assess whether it has succeeded; \textbf{Diagnose} denotes discovering what is wrong without being given the desired correction;
\textbf{No-op} denotes an explicit setting in which the input may already be correct, and the model should refrain from editing; and \textbf{Self-stop} denotes allowing the agent to determine when no further
intervention is required.
}
\label{tab:related_work}
\centering
\begin{adjustbox}{width=\textwidth}

\begin{tabular}{lcccccc}
\toprule
\textbf{Work}
& \textbf{Edit / Repair}
& \textbf{Conserve}
& \textbf{Verify}
& \textbf{Diagnose}
& \textbf{No-op}
& \textbf{Self-stop} \\
\midrule

ChartMimic~\citep{yang2025chartmimic}
& \xmark & \xmark & \xmark & \xmark & \xmark & \xmark \\

Chart2Code~\citep{tang2026charts}
& \cmark & \xmark & \xmark & \xmark & \xmark & \xmark \\

ChartEdit~\citep{zhao2025chartedit}
& \cmark & \cmark & \xmark & \xmark & \xmark & \xmark \\

ChartEditBench~\citep{kapadnis2026charteditbench}
& \cmark & \cmark & \xmark & \xmark & \xmark & \xmark \\

RealChart2Code~\citep{zhang2026realchart2code}
& \cmark & \cmark & \xmark & \xmark & \xmark & \xmark \\

VisEditBench~\citep{rahman2026viseditbench}
& \cmark & \xmark & \cmark & \xmark & \xmark & \xmark \\

MisViz~\citep{tonglet2026chart}
& \xmark & \xmark & \xmark & \cmark & \xmark & \xmark \\

MisVisFix~\citep{das2025misvisfix}
& \cmark & \xmark & \xmark & \cmark & \xmark & \xmark \\

\midrule
\textbf{Ours}
& \cmark
& \cmark
& \cmark
& \cmark
& \cmark
& \cmark \\

\bottomrule
\end{tabular}
    
\end{adjustbox}
\end{table}

\textbf{Chart generation and instruction-driven editing.}
Prior benchmarks study chart generation, reconstruction, and instruction-guided editing. ChartMimic~\citep{yang2025chartmimic} and Chart2Code~\citep{tang2026charts} evaluate chart-to-code generation and reconstruction, while ChartEdit~\citep{zhao2025chartedit}, ChartE$^3$~\citep{li2026charte}, and ChartEditVista~\citep{chen2026charteditor} focus on modifying visualizations under explicit instructions. More recent benchmarks extend editing to multi-turn or iterative settings: ChartEditBench~\citep{kapadnis2026charteditbench} studies successive user-guided modifications, RealChart2Code~\citep{zhang2026realchart2code} evaluates instructed refinement while preserving unaffected content, and VisEditBench~\citep{rahman2026viseditbench} introduces iterative execution and render-grounded refinement. Across these settings, however, the target generation or modification is specified externally rather than autonomously determined by the agent.

\textbf{Misleading visualization detection and correction.}
Complementary studies examine whether models can recognize deceptive or misleading chart designs. Misleading ChartQA~\citep{chen2025unmasking} evaluates reasoning over charts containing a taxonomy of misleading patterns, while Misviz~\citep{tonglet2026chart} focuses on detecting misleading visualizations and identifying violated design rules. MisVisFix~\citep{das2025misvisfix} goes beyond detection by providing an interactive system for detecting, explaining, and correcting misleading visualizations. 

Table~\ref{tab:related_work} summarizes the distinction between these settings and ours. Existing benchmarks generally isolate individual capabilities. Generation benchmarks reconstruct visualizations, editing benchmarks provide the desired modification, and detection benchmarks identify predefined defects. Some recent works introduce preservation or iterative verification, but these capabilities remain within a specified editing objective. In contrast, VisAudit evaluates the complete autonomous loop. The agent must diagnose whether something is wrong, determine what is wrong, repair only the necessary components while preserving correct content, verify the consequences of its own intervention, and determine when no further action is required. The inclusion of initially correct charts further tests whether the agent can recognize when intervention is unnecessary rather than assuming that every input requires repair.

% Existing work therefore largely focus on predefined tasks: detection-oriented benchmarks ask models to identify predefined problems, while editing-oriented benchmarks assume that the desired modification has already been specified. In contrast, we study autonomous visualization repair as a unified closed-loop problem in which an agent must determine whether a defect exists, infer what should be changed from multimodal evidence, repair only the necessary components, and verify from the resulting visualization whether the issue has been resolved. \zhu{I think this paragraph should be discussed together with table 1}

\section{VisAudit Benchmark}
VisAudit evaluates the ability of multimodal agents to autonomously review and repair data visualizations. Given a rendered chart together with an instruction and a configurable set of auxiliary modalities, an agent must determine whether the visualization contains defects and, when necessary, produce visualization code that generates a corrected chart. Unlike instruction-following chart editing, where the desired modification is
explicitly specified by a user, our benchmark emphasizes \emph{autonomous review}: depending on the benchmark track, the agent may need to autonomously identify unknown defects and determine appropriate repairs from multimodal evidence (including source data table, intended text summary, and visualization code), preserve aspects of the visualization that are already correct, and decide when no further modification is necessary.

\textbf{Task Formulation.}
Each benchmark instance contains an initial chart $V_0$ and a track-dependent instruction $I$. The initial chart $V_0$ is the artifact under review and may contain one or more visualization defects. In addition, the agent may receive a configurable collection of auxiliary modalities $\mathcal{M} \subseteq \{D, S, C_0\}$, where $D$ denotes the source data table, $S$ denotes a natural-language summary of the intended visualization, and $C_0$ denotes the visualization code used to generate $V_0$. While $V_0$ may be flawed, $D$ and $S$ provide correct reference information about the underlying data and intended visualization, allowing the agent to identify discrepancies between the observed chart and its intended content. The initial chart is always provided, whereas the availability of the auxiliary modalities depends on the experimental configuration.

The agent is tasked with producing visualization code $C^*$ whose rendered output $V^*$ corrects the initial chart $V_0$. To do so, the agent must use the auxiliary modalities to determine whether $V_0$ contains a defect and, when necessary, infer the appropriate correction. When $C_0$ is provided, the agent can inspect and modify the implementation that produced $V_0$; otherwise, it must generate visualization code based on $V_0$ and the available reference modalities. A successful output should execute correctly, faithfully represent the underlying data and intended message, repair any defects present in $V_0$, and preserve aspects that are already correct. %Thus, the benchmark evaluates whether an agent can use multimodal reference evidence to diagnose and repair visualization defects, rather than requiring pixel-level or code-level equivalence to a particular reference implementation.

\textbf{Interaction.} The agent operates iteratively in an execution-grounded environment: it observes the initial chart, instruction, and available auxiliary modalities; creates or modifies visualization code; executes the code; and inspects the resulting visualization and execution feedback. The agent may revise its solution over multiple turns and autonomously determines when no further intervention is necessary; evaluator outcomes are not exposed during interaction. Thus, the task jointly evaluates what intervention to perform and when to stop. Full details of the interactive environment, tool interface, and execution procedure are in Appendix~\ref{sec:interaction_environment}.

% \textbf{Agent Observations.}
% At the beginning of an instance, the agent observes the initial chart $V_0$, the track-specific instruction $I$, and the auxiliary modalities $\mathcal{M}$ available under the experimental configuration. These modalities
% provide complementary evidence for determining whether the visualization is consistent with its underlying data and intended content. During interaction, the agent may additionally observe feedback resulting from its previous actions, including the resulting visualization and execution
% information. Thus, the agent can update its diagnosis and repair based on evidence accumulated throughout the interaction. Section~\ref{sec:interaction_environment} describes the concrete environment, tools, and feedback exposed to the agent.

% \textbf{Agent Actions.}
% The agent acts on the visualization through an interactive tool-use process. It may inspect the available evidence, create or modify visualization code, execute candidate solutions, inspect the resulting feedback, and continue
% revising its solution as needed. The interaction terminates when the agent
% decides that no further intervention is necessary. Accordingly, the task requires the agent to make two coupled decisions: \emph{what intervention to perform} and \emph{when to stop}. The concrete
% tool interface, execution procedure, and stopping mechanism are specified in Section~\ref{sec:interaction_environment}.

\textbf{Benchmark Tracks.}
We define three complementary tracks that progressively increase the
amount of diagnosis and verification required from the agent.

\textbf{Track A: Diagnosed Repair.}
Track A evaluates visualization repair when the defect is already known. Every instance in this track contains a flawed visualization, and the instruction explicitly provides a textual description of the injected flaw. For example, the agent may be informed that
\emph{``the legend identifying the plotted series is missing.''} The description identifies the type of defect present in the visualization, and the agent must determine how to correct it using the available modalities and produce a valid repaired visualization. This track therefore focuses on the agent's repair capability without requiring it to first discover the defect.

\textbf{Track B: Autonomous Repair.}
Track B evaluates autonomous visualization diagnosis and repair. As in Track A, every instance contains a flawed visualization. However, the instruction does not reveal the injected defects. The agent must instead identify the defects autonomously by reasoning over the chart and the available auxiliary modalities, and then generate or modify visualization code to correct them. The distinction between Tracks A and B isolates the difficulty of autonomous diagnosis. Track A provides the defect description and therefore serves as an approximate upper-bound setting in which the diagnosis step is largely removed. Track B requires both diagnosis and repair and constitutes the primary setting of our benchmark. Comparing performance between the two tracks allows us to quantify the performance degradation caused by requiring the agent to discover visualization defects autonomously.

\textbf{Track C: Open-World Verification.}
Tracks A and B assume that every input visualization contains at least one defect. In realistic visualization-review settings, however, an agent cannot assume that every chart requires modification. Track C therefore includes both flawed and already-correct visualizations, requiring the agent to first determine whether intervention is necessary. For a flawed visualization, the agent must autonomously diagnose and repair the defects. For an already-correct visualization, the desired behavior is to recognize that the chart is correct and make no unnecessary modifications. Track C therefore extends autonomous repair to an open-world setting by additionally evaluating the agent's ability to verify correctness and avoid false-positive repairs.  

Across the three tracks, success requires producing executable visualization code that corrects genuine defects while preserving unaffected
content; Track C additionally requires recognizing when no intervention is
necessary. We formalize these criteria and metrics in Section~\ref{sec:evaluation}.

\textbf{Benchmark composition.} We construct VisAudit from aligned source visualizations by introducing controlled defects into the plotting code while retaining the original source table and summary as reference evidence. The resulting benchmark contains 1,900 flawed instances derived from 854 source charts, spanning 21 chart types and 10 flaw categories. Table~\ref{tab:taxonomy} summarizes the perturbation taxonomy. Each category introduces a targeted discrepancy between the rendered visualization and its reference evidence while otherwise preserving the source instance. Full details of the source data, perturbation procedure, and quality-control pipeline are provided in Appendix~\ref{sec:dataset}.

\begin{table}[H]

    \centering
    \caption{The ten flaw categories used to construct single-, double-, and
    triple-flaw instances.}
    \small
    \label{tab:taxonomy}
    \begin{tabular}{cc}
        \toprule
        Category & Injected discrepancy \\
        \midrule
        Wrong title & The title contradicts the source table or summary. \\
        Wrong axis label & An axis names a different quantity from the one plotted. \\
        Missing legend & The mapping identifying plotted series is absent. \\
        Swapped legend labels & Two legend labels identify each other's series. \\
        Value perturbation & A plotted value disagrees with its source value. \\
        Swapped series & Two series display each other's data. \\
        Dropped series & A series present in the source evidence is omitted. \\
        Wrong chart type & The plotting type contradicts the intended representation. \\
        Axis truncation & An axis limit clips at least one reference data value. \\
        Color contradiction & A visual element's color contradicts the summary. \\
        \bottomrule
    \end{tabular}
            
\end{table}

We include instances with one, two, or three flaws to evaluate increasingly complex repair scenarios. As summarized in Table~\ref{tab:dataset_counts}, the evaluation set contains 400 single-flaw, 900 double-flaw, and 600 triple-flaw instances across 106 flaw scenarios. Tracks A and B use the same 1,900 flawed tasks and differ only in whether the instruction reveals the injected flaws. Track C additionally includes 300 initially correct charts with source identities disjoint from the flawed set, yielding 2,200 instances in total.

\begin{table}[H]
    \centering
    \caption{Evaluation-set composition. Track C contains the same flawed
    instances as Tracks A and B, plus 300 initially correct instances.}
    \vspace{-1mm}
    \label{tab:dataset_counts}
    \begin{tabular}{cccc}
        \toprule
        Task group & Scenarios & 
        Instances per scenario & Instances \\
        \midrule
        Single flaw & 10 & 40 & 400 \\
        Two flaws & 36 & 25 & 900 \\

        Three flaws & 60 & 10 & 600 \\
        \midrule
        Flawed total & 106 & -- & 1,900 \\
        Additional correct charts & -- & -- & 300 \\
        \bottomrule
    \end{tabular}
\end{table}

\section{Evaluation Pipeline}\label{sec:evaluation}

We evaluate whether the submitted visualization code executes successfully, recovers properties altered by the injected flaws, and preserves
properties that were unaffected. Evaluation is based on structured figure properties, including titles, axis labels, series, plotted values, legends, and axis configurations. Recovery and preservation capture complementary aspects of repair: an agent may successfully correct an injected defect while simultaneously introducing unintended changes elsewhere in the visualization.

% AK: Check the recovery unit with Shicheng. The Task Success Criteria
% currently refer to injected defects; the scorer counts changed properties.
\textbf{Recovery.} Let $g$, $b$, and $s$ denote the flattened property maps extracted from the ground-truth chart, initial flawed chart, and submitted chart, respectively. For a figure property $p$, let $g_p$, $b_p$, and $s_p$ denote its value in the corresponding property map, with a distinguished missing value used when $p$ is absent. Let $\sim$ denote the property comparator: numerical values $x$ and $y$ match when $|x-y|\leq 10^{-9}\max(1,|x|,|y|)$, while text is compared after trimming and collapsing whitespace and ignoring case, and an absent property is treated as a distinct missing value. The operator $\not\sim$ is used to denote property mismatch. We define the perturbation footprint as $F \triangleq
\{p\in \operatorname{dom}(g)\cup\operatorname{dom}(b):
g_p\not\sim b_p\}$, which contains the properties changed by the benchmark perturbation and is independent of the submitted chart. For $|F|>0$, graded and binary recovery are
$$
\mathrm{Rec\text{-}S}
\triangleq
\frac{1}{|F|}
\sum_{p\in F}\mathbf{1}[s_p\sim g_p],
\qquad
\mathrm{Rec\text{-}B}
\triangleq
\mathbf{1}\left[\mathrm{Rec\text{-}S}=1\right].
$$
Thus, recovery is measured at the property level rather than the flaw-category level: a single injected flaw may alter multiple figure properties. For perturbed axis limits, a dedicated comparator checks whether the axis limits extracted from the submitted charts cover the required data extent or summary-specified range, rather than requiring them to exactly match the autoscaling padding of the ground-truth chart.

\textbf{Preservation.}
Let $K \triangleq
\bigl(\operatorname{dom}(g)\cup\operatorname{dom}(b)\bigr)\setminus F$ denote the properties unaffected by the benchmark perturbation, and let $A \triangleq
\operatorname{dom}(s)\setminus
\bigl(\operatorname{dom}(g)\cup\operatorname{dom}(b)\bigr)$ denote properties newly introduced by the submitted chart. We define graded and binary preservation as
\[
  \mathrm{Pre\text{-}S}
    \triangleq \frac{\sum_{p\in K}\mathbf{1}[s_p\sim g_p]}{|K|+|A|},
  \qquad
  \mathrm{Pre\text{-}B}\triangleq\mathbf{1}[\mathrm{Pre\text{-}S}=1].
\]
Thus, preservation rewards retaining properties unaffected by the injected flaw while penalizing unsupported properties introduced by the submission. For perturbations that modify plotted data, we exclude properties that change automatically as a consequence of the data modification (e.g., auto-scaled axis limits) from the preservation scope.

\textbf{Executability.}
Executability records whether the submitted program runs successfully and produces a figure that can be retrieved by the evaluation pipeline.

\textbf{Correct-Chart Recognition Rate.}
Correct-Chart Recognition Rate (CCR) measures whether the agent recognizes an initially correct chart and terminates without modifying its visualization code. We assess
this behavior on the 300 initially correct Track C instances. CCR complements
recovery and preservation by measuring whether an agent can determine that
intervention is unnecessary.

% AK: Agree on the operational CCR definition with Shicheng before
% adding its formula and results. The paragraph above states the objective.

\textbf{Efficiency, coverage, and aggregation.}
We report executed tool steps and model-response counts separately, along
with available token use and execution outcomes. Analyses by flaw count use
the same definitions as the overall analysis. Completed unsuccessful repairs
remain in the outcome statistics; technical failures and unstarted tasks
are reported as coverage gaps. Each metric is averaged over its defined
values with its own denominator. We report binary and graded recovery
and preservation separately.

\section{Experiments}\label{sec:experiments}
Our experiments evaluate current multimodal agents along three dimensions of autonomous visualization review. We first report overall performance across the three benchmark tracks, examining how repair changes when defect information is provided, must be inferred autonomously, or when the input may already be
correct. We then analyze performance by flaw category and number of simultaneous flaws to characterize where agents fail. Finally, we conduct ablations on the
available input modalities and interaction budget to study the roles of multimodal evidence and iterative environment interaction.

\textbf{Models and benchmark tracks.}
We use the following models to evaluate the benchmark: gemma-4-31B-it, Qwen3-VL-235B-A22B-Instruct, Ministral-3-8B-Instruct-2512, Mistral-Large-3-675B-Instruct-2512-NVFP4, claude-haiku-4-5-20251001, deepseek-v4.1-flash, and gemini-3.8-flash. We use shortened model names in tables for readability. Each model is evaluated on the same 1,900 flawed instances in Tracks A and B and
on those flawed instances plus 300 initially correct instances in Track C.
% AK: Add the complete served model identifiers and access dates to the
% appendix. Confirm the collaborators' additional model rows with Shicheng.

The default input condition provides the chart, plotting code, source table, and summary. Track A provides a textual description of the injected flaw. Track B asks the agent to diagnose and repair the chart without describing the flaws. Track C additionally requires the agent to determine whether a repair is necessary. 

\textbf{Interaction and inference.}
Agents inspect files, edit code, execute it, and view rendered charts through the shared interactive environment. In the default setting, we impose no fixed limit on the number of interaction turns. Instead, the agent autonomously determines when the review is complete and terminates the interaction by calling \texttt{finish}. This self-terminating setting evaluates not only the agent's ability to diagnose and repair a visualization, but also its ability to determine when no further intervention is necessary. During inference, we use a temperature of 0 and a maximum of 8,000 output tokens per model response. Completed
trajectories record the model identity, tool actions, termination reason, and available token usage.

% \textbf{Interaction and inference.}
% Agents inspect files, edit code, execute it, and view rendered charts through
% the shared environment. The main condition allows the agent to choose its
% interaction length and submit its assessment with \texttt{finish}.
% Provider requests use temperature 0 and at most 8,000 output tokens per
% response. Completed trajectories record the model identity, tool actions,
% termination reason, and available token usage.
% AK: Reconcile the termination description with the actual run manifests
% before finalizing this section. Discuss the stagnation rule and technical
% failure handling with Shicheng; list the agreed rules explicitly.

% \textbf{Ablation design.}
% All ablations use Gemma and Qwen on all 1,900 flawed Track-B tasks. The
% modality ablation removes code, table, or summary individually while retaining
% the chart. The interaction ablation uses all four modalities and limits the
% agent to 3, 5, or 10 executed tool actions, including \texttt{finish}. The
% full-input, self-terminating interactive setting serves as the reference for
% these comparisons.

\textbf{Evaluation.}
We evaluate agents using executability (Exec.), binary and graded recovery (Rec-B/Rec-S), and binary and graded preservation (Pre-B/Pre-S), as defined in Section~\ref{sec:evaluation}. Recovery and preservation metrics are reported as percentages. We additionally report interaction length as the number of executed tool actions (Turns) and inference cost as the mean input and output tokens per episode (In/Out). Token counts are summed over all model responses within an episode; input counts include cached tokens and may not be directly comparable across models due to tokenizer differences. For Track C, we additionally report correct-chart recognition (CCR) on the 300 initially correct charts. Results are further analyzed by flaw count and flaw category.

% \begin{table}[H]
% \centering\small
% \setlength{\tabcolsep}{3.5pt}
% \caption{Track B: Autonomous Repair on 1,900 flawed instances.}
% \label{tab:track_b_results}
% \begin{tabular}{lcccccccc}
% \toprule
% Model & Exec. & Rec-B & Rec-S & Pre-B & Pre-S & Turns & In tok. & Out tok. \\
% \midrule
% Gemma & 99.79 & 39.68 & 70.09 & 50.32 & 85.53 & 13.20 & 56.3 & 2.9 \\
% Mistral & 98.37 & 36.79 & 65.95 & 59.24 & 88.22 & 8.53 & 29.0 & 2.3 \\
% Claude & 99.89 & 30.79 & 62.49 & 69.04 & 91.70 & 9.01 & 48.5 & 2.6 \\
% Qwen & 99.53 & 27.68 & 58.55 & 66.67 & 91.13 & 9.18 & 67.0 & 2.7 \\
% Ministral & 87.74 & 25.11 & 52.21 & 37.81 & 70.06 & 10.53 & 73.1 & 3.0 \\
% DeepSeek  & 96.79 & 45.63 & 68.84 & 74.09 & 94.01 & 6.07 & 21.1 & 8.4 \\
% Gemini & 90.37 & 47.42 & 65.18 & 78.99 & 96.24 & 6.87 & 28.6 & 7.6 \\
% \bottomrule
% \end{tabular}
% \end{table}

\begin{wraptable}[9]{r}{0.65\textwidth}
\vspace{-4mm}
\centering
\scriptsize
\caption{Track B: Autonomous Repair on 1,900 flawed instances.}
\vspace{-1mm}
\label{tab:track_b_results}
\begin{tabular}{lcccccccc}
\toprule
Model & Exec. & Rec-B & Rec-S & Pre-B & Pre-S & Turns & In & Out \\
\midrule
Gemma     & 99.8 & 39.7 & 70.1 & 50.3 & 85.5 & 13.2 & 56.3 & 2.9 \\
Mistral   & 98.4 & 36.8 & 66.0 & 59.2 & 88.2 & 8.5  & 29.0 & 2.3 \\
Claude    & 99.9 & 30.8 & 62.5 & 69.0 & 91.7 & 9.0  & 48.5 & 2.6 \\
Qwen      & 99.5 & 27.7 & 58.6 & 66.7 & 91.1 & 9.2  & 67.0 & 2.7 \\
Ministral & 87.7 & 25.1 & 52.2 & 37.8 & 70.1 & 10.5 & 73.1 & 3.0 \\
DeepSeek  & 96.8 & 45.6 & 68.8 & 74.1 & 94.0 & 6.1  & 21.1 & 8.4 \\
Gemini    & 90.4 & 47.4 & 65.2 & 79.0 & 96.2 & 6.9  & 28.6 & 7.6 \\
\bottomrule
\end{tabular}
\end{wraptable}
\textbf{Autonomous repair remains challenging.}
Table~\ref{tab:track_b_results} shows that high executability does not necessarily translate into successful repair. Most models produce executable outputs in over $96\%$ of instances, yet binary recovery remains below $50\%$ for all models. The gap between binary and graded recovery is also substantial, indicating that agents often correct some, but not all, properties affected by a flaw. Recovery and preservation further reveal different model behaviors. Gemini and DeepSeek achieve the highest binary recovery ($47.4\%$ and $45.6\%$) while also attaining strong preservation, whereas Gemma achieves a relatively high recovery score but substantially lower preservation. This suggests that successful autonomous repair requires not only identifying and correcting the target flaw, but also avoiding unintended modifications to already-correct chart properties. Interaction behavior also varies considerably: on average, DeepSeek and Gemini terminate after roughly 6--7 tool actions, while Gemma uses more than 13.

\begin{wraptable}[18]{r}{0.62\textwidth}
\vspace{-4mm}
\centering
\scriptsize
\caption{Repair performance on the 1,900 flawed instances in Tracks A and C.}
\vspace{-1mm}
\label{tab:track_ac}
\begin{tabular}{lccccccc}
\toprule
Model & Track & Rec-B & Rec-S & Pre-B & Pre-S & Turns & In/Out \\
\midrule
\multirow{2}{*}{Gemma}
& A & 57.1 & 82.2 & 71.6 & 92.6 & 12.2 & 50.0/2.8 \\
& C & 35.2 & 66.8 & 37.5 & 78.5 & 13.0 & 54.1/2.6 \\
\midrule
\multirow{2}{*}{Qwen}
& A & 60.2 & 83.8 & 84.5 & 96.4 & 8.1 & 43.2/2.8 \\
& C & 26.4 & 55.9 & 66.1 & 89.9 & 9.0 & 65.2/2.7 \\
\midrule
\multirow{2}{*}{Ministral}
& A & 35.1 & 57.2 & 55.9 & 79.2 & 9.2 & 45.0/2.9 \\
& C & 26.4 & 51.9 & 45.8 & 75.3 & 9.9 & 61.7/3.1 \\
\midrule
\multirow{2}{*}{Claude}
& A & 63.1 & 86.1 & 87.5 & 97.6 & 8.6 & 43.5/2.4 \\
& C & 32.9 & 65.1 & 58.3 & 85.5 & 8.7 & 40.4/2.2 \\
\midrule
\multirow{2}{*}{Mistral}
& A & 56.7 & 80.1 & 77.8 & 95.2 & 8.1 & 26.6/2.2 \\
& C & 41.1 & 69.8 & 52.3 & 85.5 & 8.4 & 26.9/2.1 \\
\midrule
\multirow{2}{*}{DeepSeek}
& A & 63.1 & 80.9 & 91.4 & 98.1 & 5.4 & 17.7/5.4 \\
& C & 47.0 & 72.1 & 73.1 & 92.6 & 6.2 & 20.9/7.5 \\
\midrule
\multirow{2}{*}{Gemini}
& A & 68.0 & 82.1 & 97.3 & 99.3 & 6.4 & 26.1/6.3 \\
& C & 48.2 & 68.2 & 75.2 & 95.0 & 6.9 & 28.7/7.1 \\
\bottomrule
\end{tabular}
\end{wraptable}
\textbf{Effect of diagnostic supervision and open-world verification.}
Having established autonomous repair performance in Track B, we next examine how performance changes when the flaw diagnosis is provided (Track A) or when the agent must operate without knowing whether the input chart is flawed (Track C). Table~\ref{tab:track_ac} reports performance on the same 1,900 flawed instances. Providing the diagnosis in Track A substantially improves recovery across models, indicating that identifying the underlying defect is a major bottleneck in autonomous repair. For example, Qwen reaches $60.2\%$ binary recovery in Track A, compared with $27.7\%$ in Track B, while Claude reaches $63.1\%$ compared with $30.8\%$. Preservation generally improves as well, showing that an explicit diagnosis helps agents target their interventions and avoid unintended changes. On the flawed subset of Track C, recovery is generally comparable to Track B, while preservation also decreases for several models. The additional uncertainty over whether intervention is necessary therefore affects not only the decision to act but also the subsequent repair behavior. Notably, turn counts change relatively little across tracks for most models, suggesting that the performance differences arise primarily from diagnosis and decision making rather than simply from longer interaction.

\begin{wraptable}[10]{r}{0.57\textwidth}
\vspace{-4mm}
\centering
\scriptsize
\caption{Behavior on the 300 initially correct instances in Track C.}
\vspace{-1mm}
\label{tab:clean_behavior}
\begin{tabular}{lcccccc}
\toprule
Model & Exec. & CCR & Pre-B & Pre-S & Turns & In/Out \\
\midrule
Gemma     & 99.7 & 34.3 & 50.7 & 87.4 & 12.1 & 45.6/2.5 \\
Qwen      & 100.0 & 72.0 & 88.0 & 96.4 & 7.5 & 42.7/1.9 \\
Ministral & 94.7 & 33.0 & 64.7 & 83.9 & 9.9 & 54.8/2.5 \\
Claude    & 100.0 & 61.0 & 70.0 & 92.7 & 9.4 & 52.0/2.8 \\
Mistral   & 99.0 & 32.7 & 59.7 & 90.5 & 7.8 & 20.4/1.8 \\
DeepSeek  & 94.0 & 63.3 & 81.3 & 95.9 & 6.0 & 19.5/9.3 \\
Gemini    & 99.0 & 64.0 & 79.3 & 95.8 & 6.9 & 28.8/6.4 \\
\bottomrule
\end{tabular}
\end{wraptable}
\textbf{Behavior on initially correct charts.}
The preceding comparison considers only flawed instances in Track C. We further evaluate whether agents can recognize when no intervention is necessary using the 300 initially correct charts. As shown in Table~\ref{tab:clean_behavior}, CCR varies
substantially across models, ranging from $32.7\%$ to $72.0\%$. Even when the input chart is already correct, agents frequently modify it rather than terminate without intervention. Moreover, binary preservation remains below $90\%$ for all models, indicating that unnecessary interventions can alter originally correct chart properties. These findings highlight a distinct
challenge in open-world visualization review: beyond diagnosing and repairing defects, agents must also determine when to leave a visualization unchanged.
%\julian{why is Pre-B different than CCR? I would think that if the agent detects a correct chart, it would have no modifications and Pre-B=1; if it thinks the chart is incorrect, it would modify it and Pre-B=0.}

\begin{table}[H]
\centering
\scriptsize
\setlength{\tabcolsep}{2.0pt}
\caption{Full recovery (Rec-B) by flaw category across Tracks A/B/C. Each category contains 40 single-flaw instances.}
\vspace{-1mm}
\label{tab:flaw_categories}
\begin{tabular}{lccccccc}
\toprule
Flaw & Gemma & Qwen & Ministral & Claude & DeepSeek & Gemini & Mistral \\
\midrule
Axis truncation
& 95.0/85.0/85.0 & 77.5/37.5/35.0 & 45.0/47.5/47.5 & 80.0/32.5/32.5
& 92.5/70.0/65.0 & 92.5/50.0/60.0 & 40.0/67.5/72.5 \\

Color contradiction
& 62.5/52.5/37.5 & 62.5/30.0/17.5 & 22.5/12.5/10.0 & 72.5/57.5/45.0
& 65.0/65.0/62.5 & 72.5/52.5/57.5 & 47.5/25.0/37.5 \\

Dropped series
& 82.5/47.5/35.0 & 87.5/65.0/70.0 & 40.0/27.5/32.5 & 95.0/75.0/65.0
& 87.5/60.0/65.0 & 87.5/77.5/75.0 & 65.0/30.0/32.5 \\

Missing legend
& 75.0/30.0/30.0 & 75.0/15.0/20.0 & 40.0/30.0/30.0 & 77.5/27.5/40.0
& 77.5/55.0/60.0 & 75.0/45.0/47.5 & 52.5/25.0/45.0 \\

Swapped legend labels
& 70.0/50.0/42.5 & 72.5/65.0/55.0 & 50.0/32.5/32.5 & 82.5/62.5/55.0
& 77.5/55.0/67.5 & 77.5/60.0/62.5 & 57.5/65.0/65.0 \\

Swapped series
& 72.5/65.0/55.0 & 80.0/67.5/67.5 & 52.5/50.0/45.0 & 85.0/67.5/65.0
& 92.5/65.0/65.0 & 95.0/75.0/72.5 & 72.5/65.0/60.0 \\

Value perturbation
& 70.0/47.5/47.5 & 67.5/65.0/52.5 & 57.5/37.5/37.5 & 75.0/57.5/52.5
& 75.0/60.0/55.0 & 70.0/55.0/47.5 & 62.5/47.5/47.5 \\

Wrong axis label
& 75.0/70.0/70.0 & 77.5/67.5/55.0 & 52.5/52.5/50.0 & 72.5/65.0/67.5
& 70.0/70.0/75.0 & 70.0/70.0/72.5 & 45.0/62.5/62.5 \\

Wrong chart type
& 32.5/30.0/25.0 & 42.5/25.0/25.0 & 22.5/25.0/22.5 & 45.0/32.5/35.0
& 32.5/35.0/32.5 & 52.5/40.0/40.0 & 35.0/30.0/40.0 \\

Wrong title
& 100.0/92.5/95.0 & 100.0/92.5/92.5 & 77.5/80.0/85.0 & 97.5/97.5/97.5
& 90.0/90.0/95.0 & 92.5/92.5/85.0 & 77.5/95.0/100.0 \\
\bottomrule
\end{tabular}
\end{table}

\textbf{Difficulty varies substantially across flaw categories.}
Table~\ref{tab:flaw_categories} breaks down full recovery by flaw type across all three tracks. Wrong titles are consistently among the easiest flaws to repair, whereas wrong chart types remain challenging even in Track A, where the flaw is explicitly provided. This indicates that some failures arise not from identifying the defect, but from successfully executing the required repair. Comparing Tracks A and B further reveals where autonomous diagnosis becomes a bottleneck: providing the flaw yields particularly large gains for categories such as missing legends, dropped series, and color contradictions for many models, while the gap is generally smaller for readily identifiable flaws such as wrong titles and axis labels. Performance in Tracks B and C is often closer. Although Track C adds uncertainty over whether intervention is necessary, its different open-world instruction can change agent behavior, so recovery on the flawed subset is not necessarily lower than Track B for every model--flaw combination. Together, these results expose distinct limitations at different stages of autonomous review: carrying out a known repair, identifying an unspecified defect, and deciding how to act when the need for intervention is itself uncertain.

\begin{table}[H]
\centering
\caption{Recovery by number of flaws across Tracks A, B, and C. Each entry reports single/double/triple-flaw performance.}
\label{tab:recover_flaw_count}
\begin{adjustbox}{width=\textwidth}

\begin{tabular}{lcccccc}
\toprule
& \multicolumn{2}{c}{Track A}
& \multicolumn{2}{c}{Track B}
& \multicolumn{2}{c}{Track C} \\
\cmidrule(lr){2-3}\cmidrule(lr){4-5}\cmidrule(lr){6-7}
Model & Rec-B & Rec-S & Rec-B & Rec-S & Rec-B & Rec-S \\
\midrule
Gemma
& 73.5/57.6/45.5 & 82.5/82.1/82.1
& 57.0/38.3/30.2 & 68.9/69.4/72.0
& 52.2/33.8/26.0 & 65.5/66.2/68.4 \\

Qwen
& 74.2/63.1/46.3 & 82.6/84.3/83.7
& 53.0/23.0/17.8 & 62.1/57.3/58.1
& 49.0/22.2/17.5 & 56.9/55.0/56.6 \\

Ministral
& 46.0/35.0/28.0 & 53.8/57.1/59.5
& 39.5/23.6/17.8 & 50.8/51.3/54.6
& 39.2/24.8/20.2 & 51.0/50.8/54.1 \\

Claude
& 78.2/63.4/52.3 & 85.2/85.9/87.0
& 57.5/25.9/20.3 & 66.5/60.5/62.8
& 55.5/28.9/23.8 & 66.1/62.9/67.8 \\

Mistral
& 55.5/61.6/50.2 & 64.1/84.2/84.7
& 51.3/35.6/29.0 & 62.4/65.4/69.1
& 56.2/40.3/32.2 & 67.2/69.1/72.7 \\

DeepSeek
& 76.0/64.6/52.2 & 82.0/83.0/77.2
& 63.0/46.2/33.2 & 70.0/68.9/68.0
& 64.3/46.4/36.2 & 72.9/71.8/71.9 \\

Gemini
& 79.0/69.2/58.8 & 84.1/82.6/80.2
& 60.5/46.8/39.7 & 67.0/63.9/65.9
& 62.0/47.7/39.8 & 69.8/67.4/68.3 \\
\bottomrule
\end{tabular}
    
\end{adjustbox}
\end{table}

\textbf{Complete recovery degrades as flaws accumulate.}
Table~\ref{tab:recover_flaw_count} examines recovery as the number of flaws increases across all three tracks. Binary recovery generally decreases from single- to triple-flaw instances, while graded recovery remains substantially more stable. The decline in Rec-B is particularly pronounced in Tracks B and C. For example, Qwen drops from $53.0\%$ to $17.8\%$ in Track B and from $49.0\%$ to $17.5\%$ in Track C, compared with $74.2\%$ to $46.3\%$ in Track A. Similar gaps for several models suggest that identifying multiple simultaneous defects is an important bottleneck when the flaw information is not provided. In contrast, Rec-S changes much less with flaw count and can even increase. This is because Rec-S measures the fraction of affected properties recovered: adding flaws enlarges the recovery target but does not require every affected property to be repaired, and differences in flaw composition can make the average recovery fraction non-monotonic. The growing gap between Rec-B and Rec-S therefore indicates that agents often recover a substantial fraction of affected properties in multi-flaw instances while failing to recover all of them. Thus, increasing the number of flaws primarily reduces the likelihood of complete recovery rather than eliminating meaningful partial progress.

\begin{wraptable}[11]{r}{0.57\textwidth}
\vspace{-4mm}
\centering
\scriptsize
\caption{Track-B modality ablation on 1,900 instances.}
\vspace{-1mm}
\label{tab:modality-results}
\begin{tabular}{llccccc}
\toprule
Model & Omitted & Rec-B & Rec-S & Pre-B & Pre-S & Turns \\
\midrule
\multirow{4}{*}{Gemma}
& None    & 39.7 & 70.1 & 50.3 & 85.5 & 13.2 \\
& Code    &  9.4 & 38.9 &  0.9 & 36.7 & 12.5 \\
& Table   & 38.8 & 69.7 & 57.5 & 89.2 & 12.0 \\
& Summary & 13.1 & 39.5 & 61.1 & 91.5 & 13.5 \\
\cmidrule(lr){2-7}
\multirow{4}{*}{Qwen}
& None    & 27.7 & 58.6 & 66.7 & 91.1 & 9.2 \\
& Code    & 10.2 & 41.8 &  0.8 & 33.8 & 9.3 \\
& Table   & 24.5 & 55.9 & 77.7 & 94.2 & 7.8 \\
& Summary &  9.4 & 34.9 & 64.7 & 91.2 & 7.9 \\
\bottomrule
\end{tabular}
\vspace{-0.5em}
\end{wraptable}

\textbf{Different modalities play complementary roles in repair.}
Table~\ref{tab:modality-results} ablates the auxiliary inputs in Track B. Removing the source table has relatively little effect on recovery, whereas removing either the code or summary substantially reduces recovery. Without the summary, Rec-B falls to $13.1\%$ for Gemma and $9.4\%$ for Qwen, suggesting that the intended semantic content provides important evidence for identifying what should be corrected. Removing the code reduces recovery and collapses binary preservation to below $1\%$, showing its importance for targeted edits. The higher preservation observed when the table or summary is omitted should not be interpreted as improved repair: it coincides with lower recovery and is consistent with less effective interventions that leave more unaffected properties unchanged. Overall, the summary primarily helps determine \emph{what} to change, while the original code enables controlled execution of those changes.

\begin{wraptable}[11]{r}{0.52\textwidth}
\vspace{-4mm}
\centering
\scriptsize
\caption{Track-B ablation over interaction budgets.}
\vspace{-1mm}
\label{tab:budget-results}
\begin{tabular}{llcccc}
\toprule
Model & Budget & Rec-B & Rec-S & Pre-B & Pre-S \\
\midrule
\multirow{4}{*}{Gemma}
& 3   & 0.0  & 0.0  & 100.0 & 100.0 \\
& 5   & 0.0  & 0.0  & 100.0 & 100.0 \\
& 10  & 39.3 & 67.6 & 56.9  & 88.0 \\
& Unrestricted & 39.7 & 70.1 & 50.3 & 85.5 \\
\cmidrule(lr){2-6}
\multirow{4}{*}{Qwen}
& 3   & 0.0  & 0.0  & 100.0 & 100.0 \\
& 5   & 28.5 & 59.0 & 62.5  & 86.9 \\
& 10  & 27.4 & 58.0 & 65.6  & 90.1 \\
& Unrestricted & 27.7 & 58.6 & 66.7 & 91.1 \\
\bottomrule
\end{tabular}
\vspace{-0.5em}
\end{wraptable}

\textbf{Interaction exhibits diminishing returns beyond a sufficient budget.}
The experimental results above use the default setting in which the agent determines when to stop, without a fixed interaction limit. Table~\ref{tab:budget-results} evaluates sensitivity to the available interaction budget. Very small budgets can prevent agents from completing the repair workflow: neither model recovers any instance within three turns, and Gemma still achieves no recovery with five. Once a sufficient budget is available, however, additional interaction yields limited gains. Gemma reaches $39.3\%$ Rec-B with 10 turns, compared with $39.7\%$ under unrestricted interaction, while Qwen reaches comparable performance with five turns. Preservation also tends to decrease as the budget grows, indicating that additional interaction can introduce unnecessary changes without improving recovery. Overall, interaction is necessary to complete the repair process, but allowing more turns beyond a sufficient budget provides little benefit and can increase unintended modifications.

\section{Conclusions}
We introduced VisAudit, a benchmark for evaluating autonomous visualization review through diagnosis, repair, and verification. Across three tracks spanning diagnosed repair, autonomous repair, and open-world verification, our experiments show that current multimodal agents remain far from being capable of reliable autonomous review. Further analyses reveal substantial variation across flaw types and counts. VisAudit provides a foundation for developing reliable multimodal agents that can autonomously determine when to act, what to change,
and when their intervention succeeds.

\bibliography{iclr2027_conference}
\bibliographystyle{iclr2027_conference}

\appendix

\section{Dataset Construction}\label{sec:dataset}

VisAudit pairs flawed visualizations with multimodal reference evidence that captures the intended visualization. We construct these instances by introducing controlled defects into otherwise aligned visualization examples while preserving the source evidence, enabling agents to diagnose and repair defects through cross-modal reasoning. This construction also supports controlled evaluation of both defect correction and preservation of unaffected chart properties.

\textbf{Source visualizations.}
We use ChartNet's \texttt{core\_permissive} subset \citep{kondic2026chartnet}. Each ChartNet instance contains four aligned modalities: plotting code, its rendered chart, the underlying CSV table, and a natural-language summary describing the visualization. This cross-modal alignment is essential to our construction: before perturbation, the table and summary are semantically consistent with the reference chart and thus provide complementary evidence about its intended content. We execute the reference code (i.e., the initial code that renders ground truth chart) and retain the rendered chart and structured figure properties as the reference for evaluation. Starting from each aligned instance, we perturb the plotting code to introduce one or more controlled defects and execute the modified program to produce the flawed chart. Crucially, the original CSV table and summary remain unchanged. As a result, the perturbed chart may conflict with information preserved in these reference modalities, creating a setting in which an agent can detect and correct visualization defects by reasoning across modalities.

The flawed evaluation set contains 1,900 instances derived from 854 source charts
and covers 21 chart types. A source chart can appear in several instances with
different perturbations. The plotting-library labels are Seaborn (704 instances),
Matplotlib (687), and Plotly (509). Reference and perturbed code contain medians of 61 and 60 lines, respectively; summaries contain a median of 285
words. The median table contains 13 data rows and four header columns.
The chart-type distribution is in Appendix~\ref{app:dataset_statistics}.

\textbf{Perturbation taxonomy.}
We define ten flaw categories (Table~\ref{tab:taxonomy}). Each introduces a
specific discrepancy in the chart while leaving the source evidence unchanged.
For example, value perturbation changes a plotted value while preserving the
correct value in the table. Missing legend tests whether the agent recognizes that a mapping is needed to identify the plotted series, even when the text summary does not explicitly request a legend.

\textbf{Generating single and combined flaws.}
For each flaw category, we define a category-specific prompt (in Appendix~\ref{app:generation_prompts}) that instructs \texttt{Mistral-Medium-3.5-128B} to modify the reference plotting code so as to inject the corresponding defect while otherwise preserving the original visualization. We execute the modified code to obtain the flawed chart and compare its structured figure properties with those of the reference chart to verify the changes introduced by the perturbation. Axis truncation is handled separately using a deterministic transformation that modifies the axis limits and records the clipped reference value. To construct multi-flaw tasks, we sequentially apply two or three category-specific perturbations to the same reference instance. The combination policy retains 36 pair scenarios and 60 triple scenarios, excluding combinations with conflicting effects or effects that cannot be sufficiently attributed to their components. We retain records of the constituent edits and compare the final flawed chart with the reference to capture their combined effect.

\textbf{Quality control and human alignment.}
Because our flawed instances are generated through language-model edits, we validate the resulting tasks to ensure that the intended defect is successfully instantiated and that it can be identified and repaired from the available reference evidence. We first apply automatic checks that require executable plotting code and a nonempty change in the extracted figure properties. Category-specific checks further verify the intended perturbation; for example, that a reference legend-label group disappears.

Beyond these structural checks, determining whether a generated task is semantically valid and solvable requires assessing whether the unchanged source evidence supports identification and repair of the injected defect. Since manually inspecting every generated instance is impractical at our scale, we use a cross-family LLM judge panel for this assessment. The panel receives the flaw description, the summary, and the CSV table (prompt in Appendix~\ref{app:qc_prompt}) and consists of four models: \texttt{Llama-3.3-70B-Instruct}, \texttt{Qwen2.5-72B-Instruct}, \texttt{DeepSeek-V3.2}, and \texttt{gpt-oss-120b}. Each instance is evaluated by at least three models and accepted only with a majority of positive judgments. Of 10,945 generated single-flaw candidates, 8,446 (77.2\%) pass all checks. Multi-flaw instances must also preserve every component's change in the final chart and are judged per component; 3,868 of 4,370 double-flaw and 4,357 of 5,056 triple-flaw instances pass.

To validate the reliability of this LLM-based quality control, we manually inspect a random subset of 150 generated instances and compare the human assessments with the corresponding LLM-judge decisions. We manually assess whether the intended defect is present and whether the source evidence is sufficient to support its repair. We observe over $96.64\%$ agreement between human and LLM judgments, supporting the use of the LLM panel to validate the remaining instances at scale.
% Automatic checks require executable plotting code and a nonempty change in
% the extracted figure properties. Category-specific checks verify, for example,
% that a reference legend-label group disappears or that an axis limit clips a
% reference data value. A cross-family judge panel assesses whether the described
% flaw is supported by the unchanged source evidence. The panel receives the flaw
% description, summary, and first 15 lines of the CSV table. Its model families
% are Llama, Qwen, DeepSeek, and GPT-OSS. Acceptance requires judgments
% from at least three families other than the generator's family and a majority
% of positive judgments.

% The panel assesses textual and tabular evidence; rendering checks and manual
% inspection assess the instantiated chart. The evaluation release underwent
% manual quality inspection. The human-alignment protocol compares judge
% decisions with human judgments on 100--200 candidates, including accepted and
% rejected examples. Reviewers assess whether the intended flaw is present and
% whether the source evidence supports its repair. The comparison measures
% agreement and identifies false acceptances and false rejections.
% AK: Add the actual alignment sample size, reviewer count and agreement
% results. The paragraph above describes the protocol, not a completed result.

\textbf{Evaluation set.}
We sample the evaluation set to include 40 instances per single-flaw category,
25 per pair, and 10 per triple. This gives 400 single-, 900 double-, and 600
triple-flaw instances across 106 scenarios (Table~\ref{tab:dataset_counts}).
Tracks A and B use these same 1,900 instances and differ in whether the instruction
reveals the flaw categories. Flaw count, track, available modalities, and
interaction budget provide separate dimensions for studying task difficulty.

\textbf{Initially correct charts and release.}
Track C adds 300 validated, unperturbed ChartNet instances, giving 2,200 instances.
Their source identities are unique and disjoint from the 854 sources in the
flawed set. Reference checks assess executability, agreement with the source
table and the text summary. The correct and flawed
instances are mixed with a fixed seed, and the instruction does not reveal
whether a particular chart needs repair. Correct charts account for 13.6\%
of Track C.
% AK: Document the clean-selection procedure in the appendix. Check its
% chart-type distribution before deciding whether further matching is needed.

%For distribution, the flawed set is divided into two groups of 950 tasks.
%Every complete evaluation uses both groups. 
Released instance identifiers
and versioned manifests identify the task files and evaluation code.
Reference code and perturbation metadata are reserved for the evaluator and excluded from the agent's task prompt.

\subsection{Dataset statistics}\label{app:dataset_statistics}
Table~\ref{tab:chart_types} reports the source chart-type labels per task.
Counts include repeated uses of a source chart across distinct perturbations.
Code length counts physical lines; summary length counts whitespace-separated
words; CSV length counts data records after the header. Header width is
reported separately because source tables need not be rectangular.

\begin{table}[htbp]
\centering\small
\caption{Instance-weighted source chart types in the flawed and additional clean
populations. A zero denotes a source chart type absent from the clean selection.}
\label{tab:chart_types}
\begin{tabular}{lrr}
\toprule
Chart type & Flawed instances & Clean instances \\
\midrule
3D Bar Chart & 2 & 1 \\
Area Chart & 148 & 37 \\
Bar Chart & 182 & 28 \\
Box Plot & 71 & 0 \\
Bubble Chart & 107 & 14 \\
Funnel Chart & 84 & 24 \\
Heatmap & 22 & 25 \\
Histogram & 57 & 0 \\
Kernel Density Estimate Plot & 86 & 1 \\
Line Chart & 187 & 49 \\
Multi-Axes Chart & 162 & 10 \\
Pie Chart & 20 & 2 \\
Radar Chart & 52 & 19 \\
Ring Chart & 14 & 1 \\
Rose Chart & 44 & 12 \\
Scatter Plot & 157 & 31 \\
Stem Plot & 98 & 6 \\
Swarm Plot & 109 & 5 \\
Tornado Chart & 181 & 27 \\
Treemap & 21 & 3 \\
Violin Plot & 96 & 5 \\
\midrule
Total & 1,900 & 300 \\
\bottomrule
\end{tabular}
\end{table}

% AK: Add a construction-pipeline figure and representative flaw examples.
% Add the model identifiers and reproducibility details after the meeting.

\subsection{Flaw-generation prompts}\label{app:generation_prompts}
Flawed candidates are generated by \texttt{mistralai/Mistral-Medium-3.5-128B}
with temperature 0, seed 42, and at most 16,000 output tokens. Every request
uses the system prompt below and a user message that is shared across
categories except for the category instruction and the two example values in
the output contract. Axis truncation is produced by the deterministic
transformation described earlier in Section~\ref{sec:dataset} and uses no prompt. In a
multi-flaw chain, each later request receives the previously perturbed script
as \texttt{code.py}, and its user message starts with the sentence
\begin{promptbox}
\raggedright
The input already contains these intentional flaws, which must remain unchanged: [categories of the earlier stages].
\end{promptbox}
followed by an empty line. The texts are reproduced from the implementation;
the benchmark's working title is replaced by its published name, and line
breaks and em dashes are changed for typesetting only.

\textbf{System prompt.}
\begin{promptbox}
\raggedright
\footnotesize
You are generating benchmark instances for VisAudit, a research benchmark on detecting and repairing flawed data visualisations.\par
\smallskip
\#\# The data model\par
\smallskip
Every source sample consists of four mutually consistent artifacts:\par
~~* `code.py`     --- a self-contained matplotlib/\allowbreak{}seaborn script that renders one chart\par
~~* `chart.png`   --- the rendered image\par
~~* `table.csv`   --- the underlying data table\par
~~* `summary.txt` --- a natural-language description of the chart\par
\smallskip
Your job is to introduce EXACTLY ONE flaw into `code.py` so that the *rendered chart starts to lie*, while `table.csv` and `summary.txt` stay untouched and therefore stay TRUE. A downstream agent will later be shown the flawed chart together with the truthful table and summary, and must find and repair the flaw. The flaw is only usable if the truthful modalities actually contradict the new chart.\par
\smallskip
You never edit and never mention the table or the summary in the code. They are given to you only so that you can (\allowbreak{}a) choose a flaw that they contradict and (\allowbreak{}b) quote the exact evidence that proves the contradiction.\par
\smallskip
\#\# The request\par
\smallskip
Each request gives you: `sample\_\allowbreak{}id`, `category` (\allowbreak{}the flaw type you must inject, with its own instruction), `library`, `chart\_\allowbreak{}type`, the complete ground-truth script, `summary.txt`, the header plus the first rows of `table.csv`, and `candidates` --- editable targets that a static-analysis pass already located in the script. Prefer a target from `candidates`; those are verified to be plain, safely replaceable literals.\par
\smallskip
\#\# Hard constraints (\allowbreak{}violating any one of these makes the instance unusable)\par
\smallskip
1. RETURN THE COMPLETE MODIFIED SCRIPT --- every line, from the first import to the last statement. Not a diff, not a snippet, no ellipsis, never "... unchanged ...".\par
2. CHANGE AS LITTLE AS POSSIBLE. The ideal edit touches one line; most categories allow only a single string or number literal to change. Never make a second, "while I am here" improvement.\par
3. NEVER REFORMAT. Preserve the original whitespace, indentation, line breaks, blank lines, quote style (\allowbreak{}' vs "), trailing commas, comment text, import order, variable names and statement order. Every byte you did not have to touch must stay identical, so that `diff gt.py flawed.py` shows only the injected flaw.\par
4. NEVER TOUCH DATA THAT IS NOT THE TARGET OF THE FLAW. Only the literal(\allowbreak{}s) named in the category instruction may change.\par
5. KEEP THE SAVE CALL INTACT. The `savefig(\allowbreak{}...)` /\allowbreak{} `plt.savefig(\allowbreak{}...)` line (\allowbreak{}filename, dpi, bbox\_\allowbreak{}inches, everything) must stay byte-identical and must remain the last rendering action. Likewise keep `figsize`, `dpi`, `plt.subplots(\allowbreak{}...)` geometry, `tight\_\allowbreak{}layout(\allowbreak{})` and any `plt.show(\allowbreak{})` exactly as they are --- changing the canvas would make the automatic visibility check pass for the wrong reason.\par
6. THE SCRIPT MUST STILL RUN. Same Python, same pinned matplotlib/\allowbreak{}seaborn versions, no network, no filesystem access, no new imports, no new helper functions, no new variables --- unless the category explicitly asks you to insert a line. If your edit makes a keyword argument invalid for the new call, remove that keyword argument (\allowbreak{}and say so in `edit\_\allowbreak{}description`) rather than returning a script that raises.\par
7. THE SCRIPT MUST STAY DETERMINISTIC. Never introduce randomness, never change or remove a random seed, never make output depend on time or environment.\par
8. NO SELF-DOCUMENTATION. Do not add comments, markers, TODOs, "\# bug injected here", or any hint that the code was modified. The flawed script must look like code a human wrote by mistake.\par
9. THE FLAW MUST BE A PLAUSIBLE MISTAKE, not vandalism. A tired analyst could have made it. And it must be repairable: an agent holding the table and the summary must be able to state what is wrong and restore the original behaviour.\par
10. THE FLAW MUST CHANGE THE RENDERED PIXELS. If your edit would produce a pixel-identical or near-identical image, the instance is worthless --- pick a different target or decline.\par
\smallskip
\#\# Evidence contract\par
\smallskip
Together with the code you return an `evidence` object. An automatic solvability gate matches the quoted evidence against the untouched `summary.txt` and `table.csv` using case-insensitive, whitespace-normalised substring matching. Quote VERBATIM from the artifact you cite --- do not paraphrase, do not re-capitalise, do not round a number, do not drop a trailing zero. Only these keys exist; use exactly the one your category instruction prescribes:\par
\smallskip
~~* "summary": [strings that must ALL appear in summary.txt]\par
~~* "min\_\allowbreak{}summary": integer --- if set, only this many of the "summary" strings need to match\par
~~* "table": [strings that must ALL appear in table.csv, verbatim, including formatting such as trailing zeros]\par
~~* "table\_\allowbreak{}nums": [numbers that must appear in table.csv; 24 also matches "24.0"]\par
~~* "either": [strings that must appear in summary.txt OR in table.csv]\par
~~* "summary\_\allowbreak{}any\_\allowbreak{}color": true --- the weakest form, it only asserts that the summary names some colour\par
\smallskip
Evidence always describes the ORIGINAL, TRUE state that the flawed chart now contradicts --- never the value you just injected.\par
\smallskip
\#\# Declining\par
\smallskip
If the category genuinely does not fit this chart --- no suitable target, the edit would be invisible, the flaw would not be contradicted by summary or table, or the script would break --- set `applicable=\allowbreak{}false`, give a one-sentence `decline\_\allowbreak{}reason`, and leave `perturbed\_\allowbreak{}code` empty. A clean decline is far more valuable than a forced instance: declines are cheap, invalid instances poison the benchmark.\par
\smallskip
\#\# Output\par
\smallskip
Answer only through the provided structured output schema. `edits` must list every contiguous text replacement you made as an (\allowbreak{}old, new) pair, copied verbatim from the two scripts, such that applying them to the original script reproduces `perturbed\_\allowbreak{}code` exactly. This is machine-verified.\par
\smallskip
\smallskip
\#\# V2 output-contract override\par
\smallskip
For this V2 pipeline, the JSON contract at the end of the user request replaces\par
the legacy structured-output paragraph above. Return that JSON object directly;\par
there is no `edits` field. `target` is mandatory and identifies the exact\par
Figure-Object entity the postcondition must prove. For color contradictions,\par
`evidence.summary\_\allowbreak{}color\_\allowbreak{}mapping` is an additional allowed and mandatory V2 key;\par
it maps the exact summary-named entity to its exact truth color, and\par
`evidence.summary\_\allowbreak{}quote` must copy the complete, direct affirmative clause that\par
maps exactly that one entity to exactly that one named color. Decline when the\par
clause contains another chart entity or color, negation, contrast, modality, a\par
condition, or a past/\allowbreak{}former-state qualifier, and also decline when any other\par
clause in the full summary gives that target a different color or blocks its\par
truth-color claim.\par
\smallskip
Repair truth must be reconstructable from the physical summary, table, and\par
candidate code; `target` and `evidence` never supply otherwise hidden truth.\par
For wrong-title, the physical summary must contain one current, direct\par
affirmative title-to-exact-GT-text clause, that text must be the only distinct\par
title claim in the summary, and the summary must not claim the injected title.\par
For wrong-axis-label, require either a clean direct axis/\allowbreak{}quantity-to-exact-GT-\par
label summary relation or an exact unique CSV column whose complete values or\par
locators bind to that GT axis. Every label claim in the active axis/\allowbreak{}quantity\par
scope must agree with the GT label, and no competing injected-label CSV\par
relation may exist; header presence alone, an unused code string or an\par
arbitrary table cell does not count. V2 changes exactly one\par
canonical axis label with replacement mechanism (\allowbreak{}a); do not use the legacy\par
two-axis swap because the binding target identifies one axis. For wrong-chart-type,\par
require a clean direct current relation from the whole chart or target series\par
to the exact GT kind, with no competing kind claim in that scope; a different\par
non-target series may legitimately have another kind. For missing-legend, the\par
summary is a conflict check only and never reconstruction evidence. Every one\par
of at least two GT labels must instead bind through at least one of two structural\par
routes: the exact full target-axis sequence of definitely-live plotted\par
`label=\allowbreak{}`/\allowbreak{}`name=\allowbreak{}` literals occurs once in GT legend order and maps to captured GT\par
series roots, or one unique complete wide named CSV series relation occurs in\par
exact GT legend order and the captured target-axis series set/\allowbreak{}count exactly\par
equals that legend group. A subset with a hidden extra series, comment, unused\par
assignment, call inside a function/\allowbreak{}branch/\allowbreak{}condition/\allowbreak{}comprehension, or unordered\par
label group does not count. For dropped-series, the complete missing\par
label-to-value relation must bind uniquely to the physical CSV. Its GT mark kind\par
must additionally be recoverable either from a clean direct current target-series-\par
to-kind summary statement (\allowbreak{}a whole-chart statement alone is insufficient), or\par
from uniform same-axis GT kinds plus at least one definitely-live surviving\par
labeled plot call of that kind, with every such target-axis survivor call\par
resolving to that same kind. Decline on negation, contrast, modality, conditions,\par
or former-state wording.\par
\end{promptbox}

\textbf{User message.}
Square brackets mark the fields filled in per request.
\begin{promptbox}
\raggedright
\footnotesize
[category instruction]\par
\smallskip
Return one JSON object and no Markdown. It must match this contract:\par
\{\par
~~"applicable": true,\par
~~"decline\_\allowbreak{}reason": "",\par
~~"perturbed\_\allowbreak{}code": "the complete modified Python script",\par
~~"edit\_\allowbreak{}description": "one sentence",\par
~~"evidence": [category-specific evidence example],\par
~~"target": [category-specific target example]\par
\}\par
\smallskip
If the category does not apply, return applicable=\allowbreak{}false, a concise decline\_\allowbreak{}reason,\par
empty perturbed\_\allowbreak{}code, and empty evidence/\allowbreak{}target objects. Never guess target metadata:\par
it is consumed by a fail-closed semantic gate after the script renders.\par
The values above are examples of the exact keys, not values to copy. Replace every\par
example with the actual original/\allowbreak{}injected value and captured series or axis identity.\par
\smallskip
--- source\_\allowbreak{}id: [source id] \textbar{} category: [category] \textbar{} library: [library] \textbar{} chart\_\allowbreak{}type: [chart type] ---\par
\smallskip
\#\# code.py\par
```python\par
[reference plotting script]\par
```\par
\smallskip
\#\# summary.txt (\allowbreak{}truth)\par
[summary]\par
\smallskip
\#\# table.csv, first rows (\allowbreak{}truth)\par
```\par
[first 12 lines of table.csv]\par
```\par
\end{promptbox}

\textbf{Category instructions.}
Each category instruction replaces \texttt{[category instruction]} in the user
message, and the contract examples replace the two bracketed example fields.

\textbf{Wrong title.}
\begin{promptbox}
\raggedright
\footnotesize
Replace the chart's title with a plausible but factually wrong title.\par
\smallskip
Target: the single title string literal. In a matplotlib or seaborn script it is the one given in `candidates.title` (\allowbreak{}from `plt.title(\allowbreak{}...)`, `fig.suptitle(\allowbreak{}...)`, `ax.set\_\allowbreak{}title(\allowbreak{}...)` or `.set(\allowbreak{}title=\allowbreak{}...)`). A plotly script carries no static candidate, so locate the literal yourself --- it is written in one of these forms:\par
~~* `fig.update\_\allowbreak{}layout(\allowbreak{}title=\allowbreak{}'...')` or `fig.update\_\allowbreak{}layout(\allowbreak{}title\_\allowbreak{}text=\allowbreak{}'...')`;\par
~~* a dict, `fig.update\_\allowbreak{}layout(\allowbreak{}title=\allowbreak{}dict(\allowbreak{}text=\allowbreak{}'...', x=\allowbreak{}0.5, font=\allowbreak{}dict(\allowbreak{}size=\allowbreak{}18)))` or `title=\allowbreak{}\{'text': '...', 'y': 0.95, 'x': 0.5\}` --- only the `text` value is the title;\par
~~* the express call itself, `px.\textless{}kind\textgreater{}(\allowbreak{}..., title=\allowbreak{}'...')`, for bar, line, scatter, pie, box, violin, treemap, funnel and the rest.\par
\smallskip
Change that literal and nothing else. Leave the `x`, `y`, `font` and `xanchor` entries beside it as they are, and never mistake a `legend\_\allowbreak{}title\_\allowbreak{}text=\allowbreak{}`, an axis title or a `subplot\_\allowbreak{}titles=\allowbreak{}(\allowbreak{}...)` entry for the chart title.\par
\smallskip
The new title must:\par
~~* read like a real chart title --- same casing convention as the original, roughly the same length, may keep a unit or a year range in parentheses;\par
~~* be WRONG for this chart: a different metric, entity, region or time period;\par
~~* NOT be a rewording, translation, abbreviation or near-synonym of the true title, and must not reuse the true title's key nouns;\par
~~* not equal any CSV column header;\par
~~* stay at most 120 characters.\par
\smallskip
If the script sets more than one title (\allowbreak{}e.g. a suptitle plus per-axes titles), change only the one in `candidates.title` and leave the others untouched --- the remaining truthful title is part of what makes the flaw findable. A plotly `title=\allowbreak{}` dict with NO `text` key is not a second title: it only positions or styles the title that an express call already set, so leave it alone and falsify the express `title=\allowbreak{}` string instead.\par
\smallskip
Evidence: \{"summary": ["\textless{}the complete current, direct affirmative summary clause that states the ORIGINAL title character for character\textgreater{}"]\}. The clause must explicitly bind the title to that exact text (\allowbreak{}for example, "The title is ..." or "The chart is titled ..."); a loose mention of the words is insufficient. The GT text must be the only distinct title claim anywhere in the summary. The clause must not negate, hedge, contrast, condition or describe a former title, and the summary must not claim the injected or any other competing title.\par
\smallskip
Decline if there is no title string literal, if the only `title=\allowbreak{}` in the script is a styling dict without a `text` key, if the summary has no qualifying direct affirmative exact-title clause, or if it contains a second distinct title claim.\par
\end{promptbox}
\noindent Contract examples: \texttt{evidence} \texttt{\{"summary": ["The title is Original chart title."]\}}; \texttt{target} \texttt{\{"path": "axes[0].title", "truth": "original", "injected": "wrong"\}}.

\textbf{Wrong axis label.}
\begin{promptbox}
\raggedright
\footnotesize
Make an axis label describe the wrong quantity. Pick exactly ONE of these two mechanisms:\par
\smallskip
(\allowbreak{}a) Replace ONE axis-label string literal (\allowbreak{}prefer the y label; use x only if there is no y label) with a fluent, domain-plausible label that names a DIFFERENT quantity, unit or dimension. It must not be a synonym or reformulation of the true label, and must not equal any CSV header of this table. Keep the style of real chart labels (\allowbreak{}Title Case, optional unit in parentheses), at most 60 characters.\par
\smallskip
(\allowbreak{}b) If distinct x and y label literals both exist, swap the two strings between `set\_\allowbreak{}xlabel`/\allowbreak{}`set\_\allowbreak{}ylabel` (\allowbreak{}or `plt.xlabel`/\allowbreak{}`plt.ylabel`).\par
\smallskip
In matplotlib and seaborn the literal sits in `set\_\allowbreak{}xlabel`/\allowbreak{}`set\_\allowbreak{}ylabel`/\allowbreak{}`plt.xlabel`/\allowbreak{}`plt.ylabel`. A plotly script carries no static candidate; the same literal is written in one of the forms below, and both mechanisms work on all of them:\par
~~* `fig.update\_\allowbreak{}layout(\allowbreak{}xaxis\_\allowbreak{}title=\allowbreak{}'...', yaxis\_\allowbreak{}title=\allowbreak{}'...')`, also spelled `xaxis\_\allowbreak{}title\_\allowbreak{}text=\allowbreak{}` --- for (\allowbreak{}b) exchange the two strings between the two keywords;\par
~~* `fig.update\_\allowbreak{}xaxes(\allowbreak{}title\_\allowbreak{}text=\allowbreak{}'...')` and `fig.update\_\allowbreak{}yaxes(\allowbreak{}title\_\allowbreak{}text=\allowbreak{}'...')` --- for (\allowbreak{}b) exchange the two strings between the two calls;\par
~~* the nested form, `fig.update\_\allowbreak{}layout(\allowbreak{}xaxis=\allowbreak{}dict(\allowbreak{}title=\allowbreak{}'...'), yaxis=\allowbreak{}\{'title': '...'\})` --- the label is the plain string, or the `text` of a `\{'title': \{'text': '...'\}\}`; a `\{'title': \{'font': ...\}\}` holds styling only and is no target;\par
~~* an express `labels=\allowbreak{}\{'\textless{}column\textgreater{}': '\textless{}label\textgreater{}'\}` mapping --- replace the VALUE of the entry whose KEY is the column that this call passes to `x=\allowbreak{}` or `y=\allowbreak{}`, and for (\allowbreak{}b) exchange the values of the x-column entry and the y-column entry. An entry for any other column renames a legend or a hover field rather than an axis, so it is no target. Work only with entries that are already there; never add one.\par
\smallskip
If a subplot figure writes the same axis title at several `row=\allowbreak{}`/\allowbreak{}`col=\allowbreak{}` call sites, replace the string at EVERY one of them so the whole figure carries the same wrong label, and use mechanism (\allowbreak{}b) only when the figure has exactly one x title and one y title.\par
\smallskip
Change nothing else --- in particular do not touch the title, the tick labels, the tick formatter or the data.\par
\smallskip
Evidence:\par
~~* mechanism (\allowbreak{}a): \{"either": ["\textless{}the original label, verbatim, from a clean direct axis/\allowbreak{}quantity summary relation or one exact unique CSV column whose complete values/\allowbreak{}locators bind to that GT axis\textgreater{}"]\}\par
~~* mechanism (\allowbreak{}b): \{"either": ["\textless{}original x label\textgreater{}", "\textless{}original y label\textgreater{}"]\}, with each label supported by that same kind of physical relation.\par
\smallskip
Header presence alone, an unused candidate-code string and an arbitrary table body cell do not count. Every label claim in the active axis/\allowbreak{}quantity scope must agree with the exact GT label, and a competing injected-label table relation also rejects. A summary clause that negates, hedges, contrasts, conditions or describes the old label does not count either.\par
\smallskip
Decline if no axis-label string literal exists, if the script's only `labels=\allowbreak{}\{\}` entries rename columns that sit on neither `x=\allowbreak{}` nor `y=\allowbreak{}`, if the exact original label has neither a qualifying summary relation nor one exact unique, fully axis-bound CSV column, or if another claim in the active scope gives the axis a competing label.\par
\end{promptbox}
\noindent Contract examples: \texttt{evidence} \texttt{\{"either": ["The y-axis label is Original axis label."]\}}; \texttt{target} \texttt{\{"path": "axes[0].xlabel", "axis\_\allowbreak{}index": 0, "axis": "x", "truth": "original", "injected": "wrong"\}}.

\textbf{Missing legend.}
\begin{promptbox}
\raggedright
\footnotesize
Remove the chart's legend so the plotted series can no longer be identified, while the drawn data stays exactly the same. Four allowed mechanisms --- take the ONE that fits this script:\par
\smallskip
(\allowbreak{}a) Delete the ENTIRE statement that creates the legend (\allowbreak{}`plt.legend(\allowbreak{}...)`, `ax.legend(\allowbreak{}...)`, `fig.legend(\allowbreak{}...)`) --- remove the whole line including its trailing newline, leaving surrounding indentation and blank lines intact.\par
\smallskip
(\allowbreak{}b) For a seaborn axes-level call that draws an automatic legend from `hue=\allowbreak{}` (\allowbreak{}lineplot, scatterplot, barplot, boxplot, violinplot, stripplot, swarmplot, kdeplot, histplot, pointplot, countplot), insert `legend=\allowbreak{}False` as an additional keyword argument in that call. Insert it as the first keyword after the opening parenthesis and keep the rest of the call byte-identical.\par
\smallskip
(\allowbreak{}c) A plotly script that already writes `showlegend=\allowbreak{}True` inside `fig.update\_\allowbreak{}layout(\allowbreak{}...)`: flip that ONE literal to `False`. It is the smallest possible edit, so prefer it whenever the script offers it.\par
\smallskip
(\allowbreak{}d) Every other plotly script: plotly switches the legend on by itself as soon as the figure holds several traces or an express grouping column (\allowbreak{}`color=\allowbreak{}`, `symbol=\allowbreak{}`, `line\_\allowbreak{}dash=\allowbreak{}`, `pattern\_\allowbreak{}shape=\allowbreak{}`), so there is no legend statement to delete. Insert exactly ONE new line,\par
\smallskip
~~~~fig.update\_\allowbreak{}layout(\allowbreak{}showlegend=\allowbreak{}False)\par
\smallskip
immediately before the line that writes the image (\allowbreak{}`fig.write\_\allowbreak{}image(\allowbreak{}...)`, or `pio.write\_\allowbreak{}image(\allowbreak{}fig, ...)`) and at the same indentation level, spelled with the figure's own variable name.\par
\smallskip
Only that layout-level switch really removes a plotly legend. NEVER delete or rewrite a `legend=\allowbreak{}dict(\allowbreak{}...)` /\allowbreak{} `legend=\allowbreak{}\{...\}` block, a `legend\_\allowbreak{}title=\allowbreak{}` or a `legend\_\allowbreak{}title\_\allowbreak{}text=\allowbreak{}`: plotly goes on drawing the legend and merely moves it back to its default position, so every series is still named and the instance is worthless. And never set `showlegend=\allowbreak{}False` on the individual traces --- that is one edit per trace, and a single trace you miss leaves half a legend standing.\par
\smallskip
`showlegend=\allowbreak{}False` also does nothing at all in two cases, and both leave the image pixel-identical, i.e. a silent no-op: when the express `color=\allowbreak{}` column is NUMERIC, because plotly then draws a continuous colour BAR instead of a legend --- look the column up in table.csv first and decline when it holds numbers; and on a `treemap`, `sunburst`, `icicle` or `funnel\_\allowbreak{}area`, which label their segments inside the marks and carry no legend to switch off.\par
\smallskip
Mechanism (\allowbreak{}a) is only a real removal when nothing re-draws a legend automatically --- this is the\par
single most common way this category silently fails (\allowbreak{}measured: 33 of 100 shipped instances had an\par
EMPTY footprint, i.e. the rendered chart still carried a legend). Two known re-drawers:\par
~~* a seaborn call with `hue=\allowbreak{}` re-draws its own legend after your deleted `plt.legend(\allowbreak{}...)`/\allowbreak{}\par
~~~~`ax.legend(\allowbreak{}...)` line --- there you MUST use (\allowbreak{}b) (\allowbreak{}`legend=\allowbreak{}False` inside the seaborn call) and\par
~~~~ALSO delete the explicit legend statement if one exists; the pair counts as one mechanism.\par
~~* `DataFrame.plot(\allowbreak{}...)` draws a legend by default --- insert `legend=\allowbreak{}False` into that call instead\par
~~~~of only deleting a later legend line.\par
Before you answer, trace the final figure in your head: if ANY code path still produces a legend\par
object after your edit, the instance is a no-op --- pick the mechanism that switches the legend off\par
at its source, or decline.\par
\smallskip
Do NOT delete or rename the `label=\allowbreak{}` arguments, do not remove any series, do not change colours: after the edit the same marks must be drawn in the same colours --- only the key that names them is gone.\par
\smallskip
Never delete a legend statement that sits inside a loop or an `if` body when removing it would leave an empty block.\par
\smallskip
Evidence: \{"summary": ["\textless{}series name 1\textgreater{}", "\textless{}series name 2\textgreater{}", ...], "min\_\allowbreak{}summary": 2\} --- list at most three exact GT names. This evidence list does not itself establish V2 recoverability: summary text is only a conflict check and is never a summary-only reconstruction route. Every removed label must instead survive through at least one of two physical structures: (\allowbreak{}1) the exact full target-axis sequence of definitely-live top-level plotted `label=\allowbreak{}`/\allowbreak{}`name=\allowbreak{}` literals appears once in exact GT legend order and maps to captured GT series roots, or (\allowbreak{}2) one unique complete wide named CSV-series relation appears in exact GT legend order and the captured target-axis series set/\allowbreak{}count exactly equals that legend group. A subset with a hidden extra series, comment, unused assignment, arbitrary table cell, call inside a function/\allowbreak{}branch/\allowbreak{}condition/\allowbreak{}comprehension, unordered or duplicate label group, negated relation or former-state claim does not count.\par
\smallskip
Decline if the chart draws only one series, if there is no legend to remove (\allowbreak{}a plotly script that already sets `showlegend=\allowbreak{}False` has none left, and a numeric `color=\allowbreak{}` column or a treemap/\allowbreak{}sunburst has none to begin with), if fewer than two exact GT labels survive through an ordered route, or if that route covers only a subset of the target-axis series.\par
\end{promptbox}
\noindent Contract examples: \texttt{evidence} \texttt{\{"summary": ["series A", "series B"], "min\_\allowbreak{}summary": 2\}}; \texttt{target} \texttt{\{"path": "axes[0].legend.labels", "labels": ["series A", "series B"]\}}.

\textbf{Swapped legend labels.}
\begin{promptbox}
\raggedright
\footnotesize
Exchange the names of two series so that the legend labels the geometry wrongly.\par
\smallskip
A script names its series in one of several ways. Pick the ONE mechanism that fits this script and needs the smallest edit:\par
\smallskip
(\allowbreak{}a) Two plot calls carrying distinct `label=\allowbreak{}"..."` string arguments: swap the two string values between them --- after the edit, call A carries B's name and vice versa.\par
\smallskip
(\allowbreak{}b) An explicit legend label list --- `legend(\allowbreak{}["A", "B", ...])`, `legend(\allowbreak{}handles, ["A", "B"], ...)`, `legend(\allowbreak{}labels=\allowbreak{}["A", "B"])`: swap two of its string entries.\par
\smallskip
(\allowbreak{}c) Two plotly traces carrying `name=\allowbreak{}"..."` (\allowbreak{}`go.Bar(\allowbreak{}..., name=\allowbreak{}'A')`, `add\_\allowbreak{}trace(\allowbreak{}...)`, `add\_\allowbreak{}scatter(\allowbreak{}..., name=\allowbreak{}'A')`): swap the two strings.\par
\smallskip
(\allowbreak{}d) A grouping column whose values become the legend entries (\allowbreak{}`hue=\allowbreak{}`, `style=\allowbreak{}`, plotly express `color=\allowbreak{}`/\allowbreak{}`symbol=\allowbreak{}`/\allowbreak{}`line\_\allowbreak{}dash=\allowbreak{}`): swap the two category names where the script spells them out --- the two keys of the dict the frame is built from, the two occurrences in the inline column literal, the two wide column names that a `melt`/\allowbreak{}`pivot` turns into levels, or the two sides of a `rename`/\allowbreak{}`replace` mapping the script already applies. The rows keep their values and their place; they only start carrying each other's name. Swap EVERY occurrence of both names in that one column --- a half-swapped column invents a third group --- and use this mechanism only when all of them fit within about three lines; a name repeated down a long column is too many edits. Then check every other place they appear --- a later `df[df['Sector'] =\allowbreak{}=\allowbreak{} 'Oil \& Gas']` filter, a `y1=\allowbreak{}'Primary'` selection, a colour dict --- and check how colours are assigned: a palette keyed by category name follows the names by itself and stays consistent, but a POSITIONAL palette (\allowbreak{}a `palette=\allowbreak{}[...]` list, `colors[i]`) over levels that are ordered by name can move the colours as well, and a mark whose fill no longer matches its own line is a second, unwanted defect. Decline rather than ship that.\par
\smallskip
(\allowbreak{}e) The legend labels come from a grouping column you cannot rewrite in one or two literals: pass the names into the EXISTING legend call, `ax.legend(\allowbreak{}title=\allowbreak{}'Region')` becomes `ax.legend(\allowbreak{}['South', 'North'], title=\allowbreak{}'Region')`, listing every drawn series in drawing order with exactly two of them exchanged. Only do this when the drawing order is beyond doubt --- an explicit list the script loops over, an explicit `hue\_\allowbreak{}order=\allowbreak{}`/\allowbreak{}`order=\allowbreak{}`/\allowbreak{}`category\_\allowbreak{}orders=\allowbreak{}`, alphabetically sorted pivot columns, the order of the `add\_\allowbreak{}trace` calls --- and only when the legend holds nothing but those series (\allowbreak{}never when it also carries size, style or marker entries). If you would be guessing the order, decline: a guessed order can hand the labels back to the right geometry, and the instance is then worthless.\par
\smallskip
Change NOTHING else. In particular do not swap or change the colours, the data, the markers, the argument order or the order of the plot calls: if you also swapped the colours in mechanism (\allowbreak{}a), (\allowbreak{}b) or (\allowbreak{}c), the chart would become truthful again and the instance would be worthless. Only the naming may lie.\par
\smallskip
Prefer two series that are visually distinguishable (\allowbreak{}different colours or clearly different shapes) and whose names both occur in the summary or the table.\par
\smallskip
Evidence: \{"either": ["\textless{}name A\textgreater{}", "\textless{}name B\textgreater{}"]\} --- the two ORIGINAL names, verbatim.\par
\smallskip
Decline if the chart draws fewer than two named series, if no mechanism above applies without a second change, if the two series are visually indistinguishable, or if the two names occur in neither the summary nor the table.\par
\end{promptbox}
\noindent Contract examples: \texttt{evidence} \texttt{\{"either": ["series A", "series B"]\}}; \texttt{target} \texttt{\{"path": "axes[0].legend.labels", "labels": ["series A", "series B"]\}}.

\textbf{Value perturbation.}
\begin{promptbox}
\raggedright
\footnotesize
Corrupt exactly ONE number in the plotted data so that a single bar/\allowbreak{}point is wrong while the table still holds the truth.\par
\smallskip
Pick one numeric literal from an inline data series --- a list or tuple with at least five numeric elements that represents MEASURED VALUES, not years, indices, x positions or axis ticks --- and replace it with the value multiplied by one of 0.45, 1.6 or 2.25, rounded to the same precision as the original literal (\allowbreak{}integers stay integers; floats keep the same number of decimals).\par
\smallskip
Rules:\par
~~* Change exactly one element. Do not touch any other number, do not reorder the list, do not reformat or re-wrap the list.\par
~~* Prefer a value in the middle of the value range: not the largest (\allowbreak{}the axis rescales and the change can be absorbed) and never a zero.\par
~~* The change must move the mark by a visible amount --- roughly at least 10\% of the plotted value range.\par
~~* In `target`, `point\_\allowbreak{}index` is the zero-based index of the numeric source-list\par
~~~~element you changed. For a heatmap, that list must have a same-length categorical\par
~~~~label list and the exact source row must occur in table.csv; the validator binds\par
~~~~both through any pivot/\allowbreak{}reordering to the rendered cell. Decline if this mapping\par
~~~~is not exact.\par
\smallskip
Evidence: \{"table\_\allowbreak{}nums": [\textless{}the ORIGINAL value\textgreater{}]\} if the table writes the number in the same form. If the table writes it with different formatting (\allowbreak{}e.g. "0.30" vs 0.3, or with a thousands separator), use \{"table": ["\textless{}the exact token from table.csv\textgreater{}"]\} instead.\par
\smallskip
Decline if there is no inline numeric series of measured values, if the chosen value does not appear in the table, or if every candidate list is a year/\allowbreak{}index axis.\par
\end{promptbox}
\noindent Contract examples: \texttt{evidence} \texttt{\{"table\_\allowbreak{}nums": [10]\}}; \texttt{target} \texttt{\{"series\_\allowbreak{}label": "series A", "point\_\allowbreak{}index": 2, "truth\_\allowbreak{}value": 10, "injected\_\allowbreak{}value": 20\}}.

\textbf{Swapped series.}
\begin{promptbox}
\raggedright
\footnotesize
Exchange the value arrays of two series so that every point of both series lies, while the labels, colours and drawing order stay correct.\par
\smallskip
Find two inline numeric lists of the SAME length (\allowbreak{}at least 5 elements) that belong to two different plotted series and hold different values, and swap their contents between the two plot calls --- either by swapping the two list literals, or by swapping the two variable names at the point where they are passed to the plotting calls, whichever is the smaller edit.\par
\smallskip
Copy the list literals character for character, including spacing and number formatting: do not reformat, do not re-wrap across lines, do not round, do not sort.\par
\smallskip
Rules:\par
~~* Do not touch the labels, colours, markers or the order of the plot calls.\par
~~* The swap must be a real transposition (\allowbreak{}both series change), never a copy of one over the other.\par
~~* The two series must have visibly different value profiles --- otherwise nothing changes on screen.\par
\smallskip
Evidence: \{"table\_\allowbreak{}nums": [\textless{}a distinctive value of series A\textgreater{}, \textless{}a distinctive value of series B\textgreater{}]\} --- e.g. the maximum of each series, exactly as it appears in table.csv.\par
\smallskip
Decline if there are not two equal-length numeric series, if the two series carry (\allowbreak{}almost) identical values, or if their values are not verifiable in the table.\par
\end{promptbox}
\noindent Contract examples: \texttt{evidence} \texttt{\{"table\_\allowbreak{}nums": [10, 20]\}}; \texttt{target} \texttt{\{"labels": ["series A", "series B"]\}}.

\textbf{Dropped series.}
\begin{promptbox}
\raggedright
\footnotesize
Delete one entire data series from the chart while the table and the summary still describe it.\par
\smallskip
A script draws its series in one of several ways. Pick the ONE mechanism that fits this script and needs the smallest edit, and drop exactly one series with it. Leave the other series, the legend call, the axis limits, the colours and everything else untouched.\par
\smallskip
(\allowbreak{}a) A plotting call of its own: remove the COMPLETE statement --- `plt.plot(\allowbreak{}x, y2, label=\allowbreak{}'B')`, `ax.bar(\allowbreak{}...)` --- including its trailing newline. If the call's result is assigned (\allowbreak{}`markerline2, stemlines2, baseline2 =\allowbreak{} ax.stem(\allowbreak{}...)`, `bar\_\allowbreak{}plot =\allowbreak{} sns.barplot(\allowbreak{}...)`), remove that whole assignment statement, together with the statements that do nothing but style exactly those returned artists (\allowbreak{}`plt.setp(\allowbreak{}markerline2, ...)`) --- and only when nothing else reads them.\par
\smallskip
(\allowbreak{}b) A grouping keyword that splits one call into several series --- seaborn `hue=\allowbreak{}`/\allowbreak{}`style=\allowbreak{}`, plotly express `color=\allowbreak{}`/\allowbreak{}`symbol=\allowbreak{}`/\allowbreak{}`line\_\allowbreak{}dash=\allowbreak{}`: filter that one level out of the frame in place, inside the existing call --- `data=\allowbreak{}df` becomes `data=\allowbreak{}df[df['Region'] !=\allowbreak{} 'North']`, `px.line(\allowbreak{}df, ...)` becomes `px.line(\allowbreak{}df[df['Region'] !=\allowbreak{} 'North'], ...)`. Compare against the group NAME, never against a number. If the call already passes `hue\_\allowbreak{}order=\allowbreak{}`/\allowbreak{}`order=\allowbreak{}` (\allowbreak{}or a plotly `category\_\allowbreak{}orders=\allowbreak{}` list), delete that one entry from the list instead --- seaborn draws only the levels it is given.\par
\smallskip
(\allowbreak{}c) A wide frame plotted with several y columns: drop that column from the list --- `df.plot(\allowbreak{}x=\allowbreak{}'Year', y=\allowbreak{}['A', 'B', 'C'])` becomes `y=\allowbreak{}['A', 'C']`, `df[['A', 'B']].plot(\allowbreak{})` becomes `df[['A']].plot(\allowbreak{})`.\par
\smallskip
(\allowbreak{}d) A melted or long frame: filter that one category out where the frame enters the plot, exactly as in (\allowbreak{}b) --- `data=\allowbreak{}df\_\allowbreak{}long[df\_\allowbreak{}long['Metric'] =\allowbreak{}=\allowbreak{} m]` becomes `data=\allowbreak{}df\_\allowbreak{}long[(\allowbreak{}df\_\allowbreak{}long['Metric'] =\allowbreak{}=\allowbreak{} m) \& (\allowbreak{}df\_\allowbreak{}long['Sector'] !=\allowbreak{} 'Retail')]`.\par
\smallskip
(\allowbreak{}e) Series drawn by a loop over categories or columns: narrow what the loop iterates over, so that one series is never drawn --- `for s in df['Sector'].unique(\allowbreak{}):` becomes `for s in ['Retail', 'Energy']:`, `for c in df\_\allowbreak{}pivot.columns:` becomes `for c in df\_\allowbreak{}pivot.columns.drop(\allowbreak{}'Retail'):`.\par
\smallskip
(\allowbreak{}f) Several plotly traces: remove one whole `fig.add\_\allowbreak{}trace(\allowbreak{}...)` /\allowbreak{} `fig.add\_\allowbreak{}bar(\allowbreak{}...)` /\allowbreak{} `fig.add\_\allowbreak{}scatter(\allowbreak{}...)` statement.\par
\smallskip
Rules:\par
~~* Never remove a statement out of a loop body or a comprehension --- that would delete every series or leave an empty block. The one exception: a body that draws SEVERAL series per iteration (\allowbreak{}two `plt.stem(\allowbreak{}...)` calls per category) may lose one of them, as long as another drawing statement stays.\par
~~* Never remove the last remaining series --- a blank chart is rejected by the pipeline.\par
~~* The series must disappear from EVERY mark that draws it. If a line and its `fill\_\allowbreak{}between` both draw it, filter it out where the frame is built so that both lose it; if that needs more than one edit, decline.\par
~~* Never introduce a number that was not in the script: filter on the series' name, not on an index or a threshold.\par
~~* Check what runs after the plotting call. Anything that indexes the drawn artists by position (\allowbreak{}`ax.containers[1]`, `axes.flat`, `colors[i]`) must still find them, or the script raises and the instance is lost.\par
~~* Do NOT remove the series' entry from an explicit `legend(\allowbreak{}[...])` list, do NOT shorten a `palette=\allowbreak{}`/\allowbreak{}`colors=\allowbreak{}` list and do NOT delete the underlying data definition: the leftover, now-unused data and the stale legend entry are realistic symptoms of this bug and part of the repair signal.\par
\smallskip
Prefer the series whose name is clearly present in the summary or in the table.\par
\smallskip
Evidence: \{"either": ["\textless{}label of the removed series\textgreater{}"]\} --- verbatim.\par
\smallskip
The complete removed label$\rightarrow$value sequence must bind uniquely to the physical CSV. Its GT mark kind must also remain reconstructable by one of two routes: either the summary directly and currently binds the removed target series itself to that exact kind (\allowbreak{}a whole-chart statement alone is insufficient), or all captured GT series on the same axis have that kind and the flawed code retains at least one definitely-live top-level surviving plotted `label=\allowbreak{}`/\allowbreak{}`name=\allowbreak{}` call of that kind, with every such target-axis survivor call resolving to the same kind. Calls hidden in functions, branches, conditions or comprehensions do not count.\par
\smallskip
Decline if the script really draws only one series, if no mechanism above fits without a second, unrelated change, if the removed series' complete label$\rightarrow$value sequence is not uniquely bound to the physical CSV, if neither kind-recovery route holds, or if the removed series' name occurs in neither the summary nor the table.\par
\end{promptbox}
\noindent Contract examples: \texttt{evidence} \texttt{\{"either": ["series A"]\}}; \texttt{target} \texttt{\{"series\_\allowbreak{}label": "series A"\}}.

\textbf{Wrong chart type.}
\begin{promptbox}
\raggedright
\footnotesize
Render the same data with the wrong chart type.\par
\smallskip
Replace the plotting function with a signature-compatible alternative at EVERY call site of that function, keeping the positional arguments unchanged. Allowed swaps:\par
~~* matplotlib (\allowbreak{}plt./\allowbreak{}ax./\allowbreak{}axs./\allowbreak{}ax1./\allowbreak{}ax2.): bar-\textgreater{}plot, barh-\textgreater{}plot, scatter-\textgreater{}plot, plot-\textgreater{}scatter\par
~~* seaborn: barplot-\textgreater{}lineplot, lineplot-\textgreater{}scatterplot, scatterplot-\textgreater{}lineplot, boxplot-\textgreater{}violinplot, violinplot-\textgreater{}boxplot, stripplot-\textgreater{}swarmplot\par
~~* plotly express, the constructor itself: scatter-\textgreater{}line, line-\textgreater{}scatter, bar-\textgreater{}line, area-\textgreater{}line, histogram-\textgreater{}box, box-\textgreater{}violin, violin-\textgreater{}box, strip-\textgreater{}box, density\_\allowbreak{}contour-\textgreater{}scatter, density\_\allowbreak{}heatmap-\textgreater{}scatter, ecdf-\textgreater{}line, funnel-\textgreater{}bar, funnel\_\allowbreak{}area-\textgreater{}bar, pie-\textgreater{}bar, line\_\allowbreak{}polar-\textgreater{}line, scatter\_\allowbreak{}polar-\textgreater{}scatter, bar\_\allowbreak{}polar-\textgreater{}bar\par
~~* plotly graph objects, the trace class: Scatter-\textgreater{}Bar, Scattergl-\textgreater{}Bar, Bar-\textgreater{}Scatter, Histogram-\textgreater{}Bar, Box-\textgreater{}Violin, Violin-\textgreater{}Box, Funnel-\textgreater{}Bar, Waterfall-\textgreater{}Bar, Pie-\textgreater{}Bar, Scatterpolar-\textgreater{}Barpolar, Barpolar-\textgreater{}Scatterpolar\par
\smallskip
Swap ONE function or trace class and swap it at every site that calls it. Never two different ones in the same script, never a pair that is not on this list, and never a second edit alongside it.\par
\smallskip
The data, labels, title, colours and axes must stay exactly as they are. Only the function name --- and, where unavoidable, incompatible keyword arguments --- may change:\par
~~* bar-\textgreater{}plot: remove `width=\allowbreak{}`, `bottom=\allowbreak{}`, `align=\allowbreak{}`, `yerr=\allowbreak{}`, `capsize=\allowbreak{}`, `hatch=\allowbreak{}`, `edgecolor=\allowbreak{}`; keep `color=\allowbreak{}` and `label=\allowbreak{}`.\par
~~* scatter-\textgreater{}plot: remove `s=\allowbreak{}`, `cmap=\allowbreak{}`, `norm=\allowbreak{}`, `vmin=\allowbreak{}`, `vmax=\allowbreak{}`, `edgecolors=\allowbreak{}`; a SCALAR `c=\allowbreak{}'red'` becomes `color=\allowbreak{}'red'` (\allowbreak{}if `c=\allowbreak{}` holds an array, decline instead).\par
~~* plot-\textgreater{}scatter: remove `linestyle=\allowbreak{}`, `linewidth=\allowbreak{}`, `lw=\allowbreak{}`, `markersize=\allowbreak{}`, and `ms=\allowbreak{}`. Keep a compatible `marker=\allowbreak{}` such as `marker=\allowbreak{}'o'` byte-identical; do not remove it merely because the line-only size keyword must go.\par
~~* seaborn: remove `errorbar=\allowbreak{}`, `ci=\allowbreak{}`, `capsize=\allowbreak{}`, `dodge=\allowbreak{}`, `estimator=\allowbreak{}`, `orient=\allowbreak{}` when the target function does not accept them.\par
~~* plotly express --- the keywords the new constructor no longer accepts:\par
~~~~~~scatter-\textgreater{}line: `size=\allowbreak{}`, `size\_\allowbreak{}max=\allowbreak{}`, `symbol=\allowbreak{}`, `symbol\_\allowbreak{}map=\allowbreak{}`, `opacity=\allowbreak{}`, `marginal\_\allowbreak{}x=\allowbreak{}`, `marginal\_\allowbreak{}y=\allowbreak{}`, `trendline=\allowbreak{}`, `color\_\allowbreak{}continuous\_\allowbreak{}scale=\allowbreak{}`, `range\_\allowbreak{}color=\allowbreak{}`;\par
~~~~~~line-\textgreater{}scatter: `line\_\allowbreak{}group=\allowbreak{}`, `line\_\allowbreak{}shape=\allowbreak{}`, `line\_\allowbreak{}dash=\allowbreak{}`, `line\_\allowbreak{}dash\_\allowbreak{}map=\allowbreak{}`, `markers=\allowbreak{}`;\par
~~~~~~bar-\textgreater{}line: `barmode=\allowbreak{}`, `base=\allowbreak{}`, `opacity=\allowbreak{}`, `text\_\allowbreak{}auto=\allowbreak{}`, `pattern\_\allowbreak{}shape=\allowbreak{}`, `color\_\allowbreak{}continuous\_\allowbreak{}scale=\allowbreak{}`;\par
~~~~~~area-\textgreater{}line: `groupnorm=\allowbreak{}`, `pattern\_\allowbreak{}shape=\allowbreak{}`;\par
~~~~~~histogram-\textgreater{}box: `nbins=\allowbreak{}`, `histfunc=\allowbreak{}`, `histnorm=\allowbreak{}`, `barmode=\allowbreak{}`, `barnorm=\allowbreak{}`, `cumulative=\allowbreak{}`, `marginal=\allowbreak{}`, `opacity=\allowbreak{}`;\par
~~~~~~box-\textgreater{}violin: `notched=\allowbreak{}`, `boxmode=\allowbreak{}`;  violin-\textgreater{}box: `box=\allowbreak{}`, `violinmode=\allowbreak{}`;  strip-\textgreater{}box: `stripmode=\allowbreak{}`;\par
~~~~~~density\_\allowbreak{}contour-\textgreater{}scatter: `nbinsx=\allowbreak{}`, `nbinsy=\allowbreak{}`, `histfunc=\allowbreak{}`, `histnorm=\allowbreak{}`, `z=\allowbreak{}`;  ecdf-\textgreater{}line: `ecdfnorm=\allowbreak{}`, `ecdfmode=\allowbreak{}`, `lines=\allowbreak{}`, `marginal=\allowbreak{}`, `opacity=\allowbreak{}`.\par
~~* plotly graph objects --- the properties the new trace class no longer has:\par
~~~~~~Scatter-\textgreater{}Bar: `mode=\allowbreak{}`, `line=\allowbreak{}`, `line\_\allowbreak{}color=\allowbreak{}`, `fill=\allowbreak{}`, `fillcolor=\allowbreak{}`, `stackgroup=\allowbreak{}`, `connectgaps=\allowbreak{}`, `hoveron=\allowbreak{}`, and a `marker=\allowbreak{}dict(\allowbreak{}...)` that carries `size=\allowbreak{}` or `symbol=\allowbreak{}` (\allowbreak{}a `marker\_\allowbreak{}color=\allowbreak{}` or `marker=\allowbreak{}dict(\allowbreak{}color=\allowbreak{}...)` survives);\par
~~~~~~Bar-\textgreater{}Scatter: `width=\allowbreak{}`, `base=\allowbreak{}`, `offset=\allowbreak{}`, `insidetextanchor=\allowbreak{}`, `textangle=\allowbreak{}`, `constraintext=\allowbreak{}`;\par
~~~~~~Histogram-\textgreater{}Bar: `nbinsx=\allowbreak{}`, `nbinsy=\allowbreak{}`, `histfunc=\allowbreak{}`, `histnorm=\allowbreak{}`, `cumulative=\allowbreak{}`, `xbins=\allowbreak{}`;\par
~~~~~~Box-\textgreater{}Violin: `boxpoints=\allowbreak{}`, `boxmean=\allowbreak{}`, `notched=\allowbreak{}`, `whiskerwidth=\allowbreak{}`;  Violin-\textgreater{}Box: `box=\allowbreak{}`, `meanline=\allowbreak{}`, `points=\allowbreak{}`, `bandwidth=\allowbreak{}`, `side=\allowbreak{}`, `scalegroup=\allowbreak{}`;\par
~~~~~~Funnel-\textgreater{}Bar: `textinfo=\allowbreak{}`, `connector=\allowbreak{}`;  Waterfall-\textgreater{}Bar: also `measure=\allowbreak{}`, `increasing=\allowbreak{}`, `decreasing=\allowbreak{}`, `totals=\allowbreak{}`;\par
~~~~~~Scatterpolar-\textgreater{}Barpolar: `mode=\allowbreak{}`, `line=\allowbreak{}`, `line\_\allowbreak{}color=\allowbreak{}`, `fill=\allowbreak{}`, `fillcolor=\allowbreak{}`, `textposition=\allowbreak{}`;  Barpolar-\textgreater{}Scatterpolar: `width=\allowbreak{}`, `base=\allowbreak{}`, `offset=\allowbreak{}`.\par
~~* `textposition=\allowbreak{}` does not survive a swap between a bar-like and a line-like trace: `'auto'`, `'inside'` and `'outside'` are bar positions, `'top center'` and its siblings are scatter positions, and the wrong one raises. Drop the keyword whenever the swap crosses that line.\par
~~* a pie carries its data in different keywords than a bar, so rename instead of removing: `px.pie(\allowbreak{}..., names=\allowbreak{}'Region', values=\allowbreak{}'Share')` becomes `px.bar(\allowbreak{}..., x=\allowbreak{}'Region', y=\allowbreak{}'Share')`, and `go.Pie(\allowbreak{}labels=\allowbreak{}..., values=\allowbreak{}...)` becomes `go.Bar(\allowbreak{}x=\allowbreak{}..., y=\allowbreak{}...)`. Rename exactly those two, then remove the pie-only rest --- `hole=\allowbreak{}`, `textinfo=\allowbreak{}`, `insidetextorientation=\allowbreak{}`, `marker\_\allowbreak{}colors=\allowbreak{}`, `rotation=\allowbreak{}`, `sort=\allowbreak{}`, `direction=\allowbreak{}`, `pull=\allowbreak{}`. A `domain=\allowbreak{}` is different: it is what puts each pie in its own tile of a grid, and dropping it stacks every one of them into a single axes, so decline instead when the figure draws its pies that way.\par
~~* a polar express call likewise renames its data keywords: `px.line\_\allowbreak{}polar(\allowbreak{}df, r=\allowbreak{}'Score', theta=\allowbreak{}'Metric', line\_\allowbreak{}close=\allowbreak{}True)` becomes `px.line(\allowbreak{}df, x=\allowbreak{}'Metric', y=\allowbreak{}'Score')` --- `theta=\allowbreak{}` becomes `x=\allowbreak{}`, `r=\allowbreak{}` becomes `y=\allowbreak{}`, and `line\_\allowbreak{}close=\allowbreak{}`, `start\_\allowbreak{}angle=\allowbreak{}`, `direction=\allowbreak{}`, `range\_\allowbreak{}r=\allowbreak{}`, `range\_\allowbreak{}theta=\allowbreak{}`, `log\_\allowbreak{}r=\allowbreak{}` go. Never drop `r=\allowbreak{}`/\allowbreak{}`theta=\allowbreak{}` without renaming them: a call left without data draws nothing.\par
\smallskip
List every removed keyword in `edit\_\allowbreak{}description`. The resulting chart must still draw visible marks --- never produce an empty plot.\par
\smallskip
Evidence: \{"summary": ["\textless{}the complete current, direct affirmative clause that binds the whole chart or every target series to the original chart type\textgreater{}"]\}. The exact GT kind or controlled alias must occur in that relation and must be the only kind affirmed in the whole-chart/\allowbreak{}target-series scope. Negation, contrast, modality, conditions and former-state language do not count; a different non-target series may legitimately use another kind.\par
\smallskip
Decline if there is no swappable call, if the call passes its data via `x=\allowbreak{}`/\allowbreak{}`y=\allowbreak{}` keywords to `plt.plot`/\allowbreak{}`plt.scatter` (\allowbreak{}they reject those keywords), if the swap needs more keyword surgery than the rules above cover, if it would drop a data channel that the summary itself describes (\allowbreak{}a bubble `size=\allowbreak{}`, a `pattern\_\allowbreak{}shape=\allowbreak{}`) --- that would be a second flaw on top of the wrong mark --- if the summary has no qualifying direct affirmative chart-type relation, or if that scope also affirms a competing kind.\par
\end{promptbox}
\noindent Contract examples: \texttt{evidence} \texttt{\{"summary": ["Series A is shown as a bar chart."]\}}; \texttt{target} \texttt{\{"series\_\allowbreak{}labels": ["series A"], "truth\_\allowbreak{}kind": "bar", "injected\_\allowbreak{}kind": "line"\}}.

\textbf{Color contradiction.}
\begin{promptbox}
\raggedright
\footnotesize
Change the plotted colours so that they contradict a colour statement in the summary.\par
\smallskip
The summary explicitly names colours. Find which drawn element the summary attributes a colour to, and repaint that element in a clearly different colour. Two allowed mechanisms:\par
\smallskip
(\allowbreak{}a) Swap two colour literals between two elements (\allowbreak{}`color=\allowbreak{}`, `c=\allowbreak{}`, `edgecolor=\allowbreak{}`, `facecolor=\allowbreak{}`, or two entries of a `palette=\allowbreak{}[...]` /\allowbreak{} `colors=\allowbreak{}[...]` list).\par
(\allowbreak{}b) Replace a single colour literal with a strongly contrasting one.\par
\smallskip
Rules:\par
~~* Stay in the same notation: a named colour is replaced by another named colour, a hex string by another hex string.\par
~~* The new colour must be clearly distinguishable from the old one, both to a human eye and by RGB distance. Good: blue\textless{}-\textgreater{}red, green\textless{}-\textgreater{}orange, purple\textless{}-\textgreater{}yellow, brown\textless{}-\textgreater{}teal, pink\textless{}-\textgreater{}navy. Bad: blue-\textgreater{}navy, gray-\textgreater{}silver, '\#1f77b4'-\textgreater{}'\#1f78b5'.\par
~~* NEVER change a `cmap=\allowbreak{}` name and never change a seaborn palette NAME (\allowbreak{}e.g. 'deep'-\textgreater{}'muted') --- those are too subtle or may not exist in the pinned version.\par
~~* The literal you repaint must be one the drawn artist really uses (\allowbreak{}measured: 9 of 100 shipped\par
~~~~instances had an EMPTY footprint because the change never reached the render). Never repaint a\par
~~~~literal that a later `palette=\allowbreak{}`/\allowbreak{}`set\_\allowbreak{}palette`/\allowbreak{}style call overrides, a variable that is never\par
~~~~passed into a plot call, or a `color=\allowbreak{}` on a seaborn call that also carries `hue=\allowbreak{}` (\allowbreak{}seaborn then\par
~~~~takes colours from the palette and ignores `color=\allowbreak{}`). If you cannot trace the literal to a\par
~~~~visible artist, decline.\par
~~* Do not change line styles, markers, alpha, linewidth or anything else.\par
\smallskip
Evidence: \{"summary": ["\textless{}the colour word as written in the summary, lowercase\textgreater{}"], "min\_\allowbreak{}summary": 1\} --- the colour the summary asserts and the chart no longer shows. V2 additionally requires the output-contract `summary\_\allowbreak{}quote` to be the complete verbatim direct affirmative clause and `summary\_\allowbreak{}color\_\allowbreak{}mapping` to contain exactly one target entity$\rightarrow$truth-colour pair. That clause may name only the target and one truth colour, and the full summary may contain no competing or blocked colour claim for that target. The legacy \{"summary\_\allowbreak{}any\_\allowbreak{}color": true\} form is not sufficient for V2; decline rather than use it.\par
\smallskip
Decline if the summary names no attributable direct affirmative target$\rightarrow$colour relation, if another clause gives that target a different or blocked colour claim, if the named colour cannot be traced to a colour literal in the code, or if the code contains no plain-string colour literal.\par
\end{promptbox}
\noindent Contract examples: \texttt{evidence} \texttt{\{"summary\_\allowbreak{}color\_\allowbreak{}mapping": \{"series A": "blue"\}, "summary\_\allowbreak{}quote": "Series A is shown in blue."\}}; \texttt{target} \texttt{\{"series\_\allowbreak{}label": "series A", "summary\_\allowbreak{}color": "blue", "injected\_\allowbreak{}color": "red"\}}.

\subsection{Quality-control judge prompt}\label{app:qc_prompt}
All four judge models receive the same system prompt and user message, with
temperature 0 and at most 700 output tokens. A judgment is positive when the
returned \texttt{detectable} field is true; an answer without a parseable
\texttt{detectable} field is requested up to three times and otherwise yields
no judgment. For a multi-flaw task, each component flaw is judged separately
with its own edit description. The texts are reproduced from the
implementation with the same substitutions as above.

\textbf{System prompt.}
\begin{promptbox}
\raggedright
\footnotesize
You are the quality gate of VisAudit, a benchmark of deliberately flawed charts.\par
\smallskip
Each benchmark instance was built by taking a correct chart and injecting exactly one flaw into\par
its plotting code. The chart's underlying table (\allowbreak{}table.csv) and its text summary (\allowbreak{}summary.txt)\par
were NOT touched, so they still describe the CORRECT chart. A benchmark agent is later shown the\par
flawed chart plus that truthful table and summary, and must notice that something is wrong.\par
\smallskip
Your single question: given only this summary and this table, could an agent notice this flaw?\par
\smallskip
An instance is only fair if the answer is yes. Judge the two cases separately:\par
\smallskip
~~* EXPLICIT: the evidence names the changed element literally. If the summary says\par
~~~~'the title reads "Sales by Month"' and the title was changed, that is explicit and\par
~~~~the contradiction is direct.\par
~~* SEMANTIC: the evidence does not name the element, but the meaning still contradicts\par
~~~~the flaw. If the summary describes a chart of expenditure percentages over time and\par
~~~~the y-axis label was changed to "Total Enrollment", the summary never quotes the axis\par
~~~~label, yet the contradiction is unmistakable. This still counts as detectable.\par
\smallskip
Be strict about the opposite case: if the summary is generic enough that the flawed version\par
would read as perfectly consistent with it, the flaw is NOT detectable and the instance must\par
be rejected. Example: a legend is removed, but the summary never names the individual series,\par
so nobody could reconstruct which series is which.\par
\smallskip
Ignore whether the flaw is visually obvious. Judge only whether the given text and numbers\par
contradict it. Quote the deciding fragment verbatim.\par
\smallskip
Respond with a single JSON object and nothing else:\par
\{"detectable": true or false,\par
~"basis": "summary" \textbar{} "table" \textbar{} "both" \textbar{} "none",\par
~"evidence\_\allowbreak{}quote": "the exact fragment that exposes the flaw, empty if not detectable",\par
~"match\_\allowbreak{}type": "explicit" \textbar{} "semantic",\par
~"reasoning": "one sentence"\}\par
\end{promptbox}

\textbf{User message.}
\begin{promptbox}
\raggedright
\footnotesize
\#\# The injected flaw\par
category: [category]\par
what was changed: [edit description returned by the generator]\par
\smallskip
\#\# summary.txt (\allowbreak{}untouched, describes the CORRECT chart)\par
[summary]\par
\smallskip
\#\# table.csv, first rows (\allowbreak{}untouched)\par
```\par
[first 15 lines of table.csv]\par
```\par
\smallskip
Could an agent notice this flaw from the summary and table above?\par
\end{promptbox}

\section{Interactive Environment}\label{sec:interaction_environment}
VisAudit places each agent in an execution-grounded loop. At the start of an episode, the environment copies
the instance into a fresh writable workspace containing only the modalities
available in that condition; evaluator metadata and reference solutions remain
outside the workspace. The agent receives the track-specific instruction, a
workspace listing, and six provider-neutral tools:
\texttt{execute\_python}, \texttt{read\_file}, \texttt{write\_file},
\texttt{list\_files}, \texttt{view\_image}, and \texttt{finish}. We disable
parallel tool calls and execute one action at a time, yielding an auditable
sequence of observations and interventions. The fixed system and task prompts
are reported in Appendix~\ref{app:prompts}.

 \textbf{Stateful execution.}
  For each episode, the runner starts a Python worker subprocess in the copied
  task workspace. It sends each code block to the worker and receives a structured
  execution result through a separate process channel. The worker reuses the same
  Python namespace across turns, so variables, imports, and open figures persist
  while the agent iteratively inspects and revises its chart. Rendering uses the
  headless Matplotlib \texttt{Agg} backend, and file-tool paths are checked against
  the episode workspace. Each execution has a timeout enforced by the worker and
  the parent process; if the worker becomes unresponsive, the parent restarts it
  and reports that its in-memory state was lost.

\textbf{Multimodal feedback and stopping.}
An execution returns captured standard output, standard error, warnings, a
traceback when applicable, and the files and images it changed. Images are not
automatically returned after execution: the agent must invoke
\texttt{view\_image}, which returns the actual image bytes, to visually inspect
its render. Execution failures are observations rather than terminal events,
so recovery remains part of the task. Crucially, verifier outcomes are never
shown to the agent and never trigger early stopping. Under the main
\texttt{abc-uncapped-v1} protocol, there is no total tool-step or episode
wall-time limit; the agent decides when to terminate by calling
\texttt{finish}, while finite request, kernel-execution, and capture timeouts
remain as infrastructure safeguards. A response without a valid tool call is
retried twice; repeated failure terminates under a separately recorded guard.
Budget ablations instead cap executed tool actions, including \texttt{finish}.
For every episode, we retain the final workspace, termination origin, token
usage, and complete action--observation trajectory.

\subsection{Agent prompts and tool interface}\label{app:prompts}
We use a fixed system prompt across tracks and vary only the task instruction.
The following text is reproduced from the benchmark implementation; line
breaks are changed for typesetting only.

\textbf{System prompt.}
\begin{promptbox}
\raggedright
You are an autonomous data-visualization agent working in a sandboxed Python
workspace. Solve the task by calling tools, one call per turn.

The Python kernel is persistent: variables, imports and open figures survive
across calls, so you can build up state instead of re-running everything. Only
files inside the workspace exist, and there is no network access---no
downloading data, no installing packages.

Look at your own output. After rendering a chart, view the image and check it
against what the task asked for before you move on: a script that exits cleanly
can still draw the wrong chart, mislabel an axis, or silently drop a series.

Call finish when the task is complete, or when you are blocked and no further
tool call would help. In that final message, say what you did and what you
verified visually---and if something is unfinished, say plainly what and why.
\end{promptbox}

\textbf{Track B: Autonomous Repair.}
\begin{promptbox}
\raggedright
chart.py builds a chart from the data in table.csv and saves it as output.png.
The chart it currently produces is not faithful to that data: at least one thing
about it contradicts what table.csv contains or what summary.txt says the chart
shows. chart\_current.png is how the chart looks right now.

Find what is wrong, fix chart.py, and run it so that output.png shows a chart
that is faithful to the data. Change only what is actually wrong and leave the
rest of the chart as it is. Call finish when you are done.
\end{promptbox}

\textbf{Track A: Diagnosed Repair.}
Track A uses the Track-B instruction followed by one of the following exact
suffixes. For a single flaw, the suffix is
\begin{promptbox}
\raggedright
Diagnosis: the flaw is that [diagnosis]. Repair exactly that.
\end{promptbox}
For a $k$-flaw instance, it is
\begin{promptbox}
\raggedright
Diagnosis: the chart carries $k$ flaws: (1) [diagnosis 1]; \ldots;
($k$) [diagnosis $k$]. Repair exactly those.
\end{promptbox}
The bracketed fields are instantiated with the category descriptions in
Table~\ref{tab:track_a_diagnoses}; brackets are not included in the prompt.

\begin{table}[htbp]
\centering\small
\caption{Diagnosis strings appended to the Track-A task instruction.}
\label{tab:track_a_diagnoses}
\begin{tabular}{p{0.27\linewidth}p{0.65\linewidth}}
\toprule
Flaw category & Diagnosis string \\
\midrule
Wrong title & the chart title contradicts the provided data or summary \\
Wrong axis label & an axis label names a quantity other than the one plotted \\
Missing legend & the legend that identifies the plotted series is missing \\
Swapped legend labels & two legend entries are attached to each other's series \\
Value perturbation & one plotted value does not match its number in table.csv \\
Swapped series & two series plot each other's data \\
Dropped series & a series described in the data is missing from the chart \\
Wrong chart type & the chart or a series uses a chart type that contradicts summary.txt \\
Axis truncation & a value axis is cut so that some plotted data is not visible \\
Color contradiction & a plotted element's color contradicts the summary \\
\bottomrule
\end{tabular}
\end{table}

\textbf{Track C: Open-World Verification.}
The same instruction is used for both flawed and initially correct instances,
so the prompt does not reveal their status.
\begin{promptbox}
\raggedright
chart.py builds a chart from the data in table.csv and saves it as output.png;
chart\_current.png is how it looks right now. The chart may be faithful to
table.csv and summary.txt, or one or more things about it may contradict them.
Your first job is to decide which.

If the chart is faithful, do not edit chart.py: run it once so that output.png
exists and call finish. If something is wrong, fix chart.py, change only what is
actually wrong, run it so that output.png shows a faithful chart, and call
finish.
\end{promptbox}

\textbf{Tool interface.}
The opening observation also lists the available workspace files and the six
tools. \texttt{execute\_python} runs source code in the persistent kernel;
\texttt{read\_file}, \texttt{write\_file}, and \texttt{list\_files} expose
workspace-local file operations; \texttt{view\_image} returns a rendered file as
a multimodal image observation; and \texttt{finish} ends the episode. The
modality ablations use the same instruction generator, which names only files
actually present in the workspace. Tool schemas and generated task prompts are
fixed in the released protocol and identified by checksums.

\section{Usage of LLMs}
In this paper, we use LLMs to polish the writing and assist with the dataset generation. Specifically, we use a Mistral model to generate flawed data (details are in Appendix \ref{sec:dataset}).

\end{document}